\documentclass[10pt,letterpaper]{article}
\usepackage[mode=buildnew, subpreambles=true]{standalone}
\usepackage[top=0.85in,left=2.75in,footskip=0.75in]{geometry}

\usepackage{amsmath,amssymb}

\usepackage{changepage}

\usepackage[utf8x]{inputenc}

\usepackage{textcomp,marvosym}

\usepackage[noadjust]{cite} %

\usepackage{nameref,hyperref}

\usepackage[right]{lineno}

\usepackage{microtype}
\DisableLigatures[f]{encoding = *, family = * }

\usepackage[table]{xcolor}

\usepackage{array}

\newcolumntype{+}{!{\vrule width 2pt}}

\newlength\savedwidth

\usepackage[english]{babel}

\usepackage[aboveskip=1pt,
labelfont=bf,
labelsep=period,
singlelinecheck=off]{caption}

\addto\captionsenglish{} %

\makeatletter
\renewcommand{\@biblabel}[1]{\quad#1.}
\makeatother

\usepackage{lastpage,fancyhdr,graphicx}
\usepackage{epstopdf}
\fancyheadoffset[L]{2.25in}
\fancyfootoffset[L]{2.25in}
\usepackage{bbm}

\usepackage{doi}
\usepackage{mathrsfs}

\usepackage{float}

\usepackage{soul}

\usepackage[none]{hyphenat}

\usepackage{amsthm}

\theoremstyle{definition}
\newtheorem{definition}{Definition}[section]

\theoremstyle{remark}
\newtheorem{remark}{Remark}

\usepackage{mathtools}

\usepackage[boxed, noend,vlined]{algorithm2e}
\SetAlgoCaptionSeparator{.}

\graphicspath{{tikz-standalones}}

\usepackage{tikz}
\usepackage{pgfplots}
\usepackage{pgfplotstable}
\usetikzlibrary{patterns,arrows,shapes,shapes.arrows}
\usetikzlibrary{positioning}
\pgfplotsset{compat=newest}

\usepackage{todonotes}

\usepackage{longtable}
\usepackage{booktabs}
\usepackage{multirow}
\usepackage{tabularx}
\usepackage{xltabular}
\usepackage{pdflscape}

\usepackage{siunitx}
\usepackage[inline]{enumitem}

\usepackage{caption}
\usepackage{subcaption}

\newcommand{\ie}{i.\,e.,\ }

\newcommand{\eg}{e.\,g.,\ }
\newcommand{\egcite}{e.\,g.}

\newcommand{\iid}{independent and identically distributed }%

\newcommand{\introduceterm}[1]{{\emph{#1}}} %
\newcommand{\algoformat}[1]{\texttt{#1}}
\newcommand{\instanceformat}[1]{\texttt{#1}}

\newcommand{\R}{\mathbb{R}} %
\newcommand{\Rpos}{\R^{+}}  %
\newcommand{\e}{\mathrm{e}} %
\newcommand{\dd}{\mathrm{d}}%
\newcommand{\Indikator}[1]{\mathbbm{1}_{#1}}

\newcommand{\set}[1]{\{ #1 \}}
\newcommand{\Set}[1]{\big\{ #1 \big\}}

\newcommand{\Setdescr}[3][\bigm\vert]{\Set{ #2 #1 #3 }}

\newcommand{\complexityclassformat}[1]{\ensuremath{\mathsf{#1}}}
\newcommand{\NP}{\complexityclassformat{NP}}

\DeclareMathOperator{\Probop}{\mathbb{P}}

\newcommand{\prob}[2][]{\Probop_{#1} \left[ #2 \right]}

\newcommand{\integralLow}[4][x]{\int_{#2}^{#3}#4 \, \dd #1}

\newcommand{\refsec}[1]{Section~\ref{#1}}
\newcommand{\refSec}[1]{Section~\ref{#1}}

\newcommand{\figuretext}{Fig}
\newcommand{\reffig}[1]{\figuretext~\ref{#1}} %
\newcommand{\refFig}[1]{\figuretext~\ref{#1}}

\newcommand{\reftab}[1]{Table~\ref{#1}}

\newcommand{\refAlg}[1]{Algorithm~\ref{#1}}

\newcommand{\form}{\ensuremath{\mathscr{F}}}

\newcommand{\varia}{\ensuremath{x_1}}     %
\newcommand{\varib}{\ensuremath{x_2}}     %
\newcommand{\varic}{\ensuremath{x_3}}     %
\newcommand{\varis}{\ensuremath{x_4}}     %
\newcommand{\varit}{\ensuremath{x_5}}     %
\newcommand{\variu}{\ensuremath{x_6}}     %
\newcommand{\variv}{\ensuremath{x_7}}     %
\newcommand{\variw}{\ensuremath{x_8}}     %
\newcommand{\varix}{\ensuremath{x_9}}     %
\newcommand{\variy}{\ensuremath{x_{10}}}  %

\newcommand{\AllClauses}{\ensuremath{\mathbb{L}}}
\newcommand{\Clauses}{\ensuremath{L}}

\newcommand{\Alfa}{\algoformat{Alfa}}
\newcommand{\GlucoseProject}{\algoformat{Glucose}}
\newcommand{\GlucoseThree}{\algoformat{Glucose}~\texttt{3.0}}
\newcommand{\GlucoseFour}{\algoformat{Glucose}~\texttt{4.1}}
\newcommand{\MiniSAT}{\algoformat{MiniSAT}}
\newcommand{\Algorithm}{\algoformat{CDCLSolver}}

\newcommand{\Geo}[1]{\ensuremath{\text{Geo}\left(#1\right)}}

\newcommand{\vertexstd}{\ensuremath{u}}
\newcommand{\vertexalt}{\ensuremath{v}}

\newcommand{\GapSAT}{\algoformat{GapSAT}}

\newcommand{\numberofcomponents}{\ensuremath{N}}

\usepackage{bold-extra}

\begin{document}
\vspace*{0.2in}

\begin{flushleft}
{\Large
\textbf\newline{Too much information: why CDCL solvers need to forget learned clauses}
\newline
}
\\ %
Tom Krüger\textsuperscript{1\Yinyang*},
Jan-Hendrik Lorenz\textsuperscript{1\Yinyang},
Florian Wörz\textsuperscript{1\Yinyang*} %
\\
\bigskip
\textbf{1} Institute of Theoretical Computer Science, Universität Ulm, 89069 Ulm, Germany
\\
\bigskip

\Yinyang All authors contributed equally to this work and are sorted alphabetically. %

* Corresponding authors: tom.krueger@uni-ulm.de, florian.woerz@uni-ulm.de

\end{flushleft}
\section*{Abstract}

Conflict-driven clause learning (CDCL) is a remarkably successful paradigm for solving the satisfiability problem of propositional logic. Instead of a simple depth-first back\-track\-ing approach, this kind of solver learns the reason behind occurring conflicts in the form of additional clauses. However, despite the enormous success of CDCL solvers, there is still only a limited understanding of what influences the performance of these solvers in what way.

Considering different measures, this paper demonstrates, quite surprisingly, that clause learning (without being able to get rid of some clauses) can not only help the solver but can oftentimes deteriorate the solution process dramatically. By conducting extensive empirical analysis, we furthermore find that the runtime distributions of CDCL solvers are multimodal. This multimodality can be seen as a reason for the deterioration phenomenon described above. Simultaneously, it also gives an indication of why clause learning \emph{in combination with} clause deletion is virtually the de facto standard of SAT solving, in spite of this phenomenon. As a final contribution, we show that Weibull mixture distributions can accurately describe the multimodal distributions. Thus, adding new clauses to a base instance has an inherent effect of making runtimes long-tailed. This insight provides an explanation as to why the technique of forgetting clauses is useful in CDCL solvers apart from the optimization of unit propagation speed.


\section{Introduction}

Since their inception in the mid-90s~\cite{MS99GRASP, MMZZM01Chaff}, CDCL solvers have proven enormously successful in solving the satisfiability problem of propositional logic (SAT). As a case in point, we refer to the annual SAT Competition\footnote{The goal of the annual SAT Competition is to promote further improvements in the field of SAT solving by hosting a competitive event where researchers can present their newest implementation work. All submitted solvers are put up against each other to solve a pool of instances. The fastest solver wins. We refer to \url{http://www.satcompetition.org} for more detailed information.}: CDCL solvers won several of the last competitions. 
In many combinatorial fields, applied problems are nowadays even solved by reducing the problem to a SAT instance and invoking a CDCL solver (see \egcite~\cite{SLM21CDCL}), despite the $\NP$-completeness of the SAT problem.\footnote{SAT solvers have also been used in the last few years to generate computerized proofs of long-standing open problems. We refer to \cite{HeuleKM16} for a solution of the Boolean Pythagorean Triples problem, \cite{Heule18} for Schur number five, or \cite{BrakensiekHMN20} for the resolution of Keller's Conjecture with SAT solvers.}

The abbreviation CDCL stands for \introduceterm{conflict-driven clause learning}. The eponymous component of CDCL is \introduceterm{clause learning}, a mechanism that can enhance the simple exhaustive exploration of the search tree for possible satisfying assignments by learning from mistakes made and avoiding these in the future. During its execution, the solver learns additional clauses containing this learned information (we refer to \refSec{sec:cdcl} for an introduction to CDCL solvers). This enables the solver to prune the search tree and avoid re-exploring similar parts.

In addition to a clause learning scheme, many modern CDCL solvers also employ a technique to delete some of the new clauses from time to time when they were deemed not helpful by the solver. However, it is still largely a mystery whether this deletion process is only used to keep computation times low by having a manageable clause database or if there is some theoretical benefit to deleting clauses. 

Although both techniques, clause learning and clause deletion, are routinely employed in modern CDCL solvers, the theoretical underpinnings of the interactions between the two techniques are not entirely understood~\cite{FrancoM21}.

\subsection{Our contribution}

To study the effect of learned clauses on CDCL solvers, we let \GlucoseFour{} (see~\cite{audemard2009glucose, AudemardS09}), a leading CDCL solver, first learn the set~$\AllClauses$ of all conflict clauses it encounters until a solution of a given instance~$\form$ is found.\footnote{We have chosen a diverse set of satisfiable as well as unsatisfiable instances. We refer to Section~\ref{sec:experiments}.}
In a second step, we generate a multitude of different sets~$L$, where each~$L$ is a randomly sampled subset of~$\AllClauses$. Finally, we call~\GlucoseFour{} on the extended instance~$\form \cup L$. This model can be thought of in the following way: we simulate that the solver learns the clauses~$\AllClauses$ and then aggressively deletes some of the learned clauses (only keeping the ones in $L$) and forgets its current assignment (we refer to this as a ``\textbf{reset}'').

Since CDCL is nowadays the leading paradigm of successful SAT solvers, one is tempted to conjecture that clause learning is always helpful. However, using the described modification process, we demonstrate in this paper that one must be careful about this assumption. More specifically, we show that there are a surprising number of instances where the mean runtime of the extended instances is dramatically worse than the runtime on the original instance. This also holds, using runtime-independent measures like the number of conflicts or propagations that occurred towards a solution. This observation shows that the observed deterioration phenomenon cannot solely be explained by a decrease in the unit propagation speed. In particular, we have designed our sampling process in such a way that the size of the formula does not increase too much, and unit propagations can be performed at a reasonable cost.

Furthermore, the performance decrease is so substantial that it cannot be explained by pure chance. This motivates the study of the runtime distribution of extended instances to shed light on the question of what influence learned clauses have on CDCL solvers.
Focusing on the runtime distribution, we obtain as our next result that the runtime distribution of \GlucoseFour{} is multimodal.
The observed multimodality contrasts the recently obtained result that the runtime distribution of stochastic local search (SLS) SAT solvers can be described with a single distribution (namely, a lognormal distribution)~\cite{WL21Evidence}.

We continue our study to determine what kind of distribution type can be used to describe this multimodal data. We demonstrate that the runtimes of \GlucoseFour{} are mixed Weibull distributed by conducting various statistical analyses.
These distributions possess the long-tailed property for a specific parameter range, which can lead to exceedingly long runtimes.
We also verify this observation with the \GlucoseFour{} solver together with Chanseok Oh's deletion strategy %
and with \MiniSAT{}~\cite{een2003extensible}, a minimalist CDCL solver.
This leads to a better understanding of the usefulness of clause deletion techniques in CDCL solvers.

\subsection{Related work}
\label{sec:RelatedWork}

This section gives a brief overview of related works in both the study of runtime distributions and the research on CDCL solvers.

\paragraph{Studying runtime distributions of algorithms.}

In previous works, the runtime distributions of algorithms were studied.
In~\cite{FRV97SummarizingCSPHardness}, the authors presented empirical evidence for the fact that the distribution of the effort (more precisely, the number of consistency checks) required for backtracking algorithms to solve constraint satisfaction problems randomly generated at the 50\,\% satisfiable point can be approximated by the Weibull distribution (in the satisfiable case) and the lognormal distribution (in the unsatisfiable case).
Later, these results were extended to a wider region around the 50\,\% satisfiable point~\cite{RF97StatisticalAnalysis}.
In~\cite{GSCK00HeavyTailedPhenomena}, the cost profiles of combinatorial search procedures were studied. 
The authors showed that Pareto-Lévy type heavy-tails often characterize the distributions and empirically demonstrated how rapid randomized restarts can effectively eliminate heavy-tail behavior.

In the paper~\cite{WL21Evidence}, the hardness distributions of several SLS SAT solvers on logically equivalent modifications~of~a base instance were studied.
The authors included different instance generation models to rule out any influence of the model. Introducing the procedure~\Alfa{} that we adapt to CDCL solvers in our work, the paper found that lognormal distributions characterize this hardness distribution perfectly.
The approach of~\cite{WL21Evidence} lends itself to the analysis of existing~SLS-CDCL hybrid solvers, like \GapSAT~\cite{LW20OnTheEffectOfLearnedClauses}.
The advantage of the approach studied in~\cite{WL21Evidence} is that the conducted work is not lost in the case~of a restart: only the logically equivalent instance could be changed while keeping the current assignment.
The paper~\cite{ATCTC13UsingSequentialRuntimeDistributions} studied the solvers~\algoformat{Sparrow} and~\algoformat{CCASAT} and found that for randomly generated instances, the lognormal distribution is a good fit for the runtime distributions. This study was performed on the domains of randomly generated and crafted instances.

Barrero et al.~\cite{BMCR1515OnTheStatisticalDistribution} observed empirical evidence suggesting lognormally distributed runtimes in several types of population-based algorithms like evolutionary and genetic algorithms.

\paragraph{Experimental studies on CDCL solvers.}

The reason behind the fact that CDCL algorithms also incorporate a mechanism to delete (subsets of the) learned clauses from time to time was explained by Mitchell in~\cite{Mitchell05}:
Even when sufficient memory is available, the time required to perform unit propagation becomes impractical for extensive clause sets, thus reducing the solver's performance.
Audemard and Simon~\cite{AudemardS09} observed that despite this phenomenon, deleting too many learned clauses can break the learning benefit. Thus, many CDCL solvers let the maximum number of learned clauses grow exponentially. The paper~\cite{AudemardS09} lead to the development of the \GlucoseProject{} solver using ``aggressive clause deletion'' together with the ``Literals Block Distance (LBD)'' measure. Some solvers also incorporate a dynamic clause management policy, allowing the solver to \emph{freeze} some learned clauses for later use instead of deleting them~\cite{AudemardLMS11}.
In~\cite{KokkalaN20}, CDCL solver heuristics such as restarts and clause database management frameworks were analyzed by studying the resolution proofs produced by the solvers.

\paragraph{Theoretical studies on CDCL solvers.}

While our study is purely empirical, it is interesting nevertheless to mention a selection of papers that study CDCL from a theoretical perspective.
The field of proof complexity aims at gaining a theoretical understanding of the reasoning power of different proof systems. 
It is well-known that an implicit CDCL run can be interpreted as resolution proof.
In the opposite direction, there has been a line of research~\cite{BeameKS04, HertelBPG08, BussHJ08} investigating the proof-theoretic strength of~CDCL. This research culminated in papers proving that CDCL (with non-deterministic variable decisions) can efficiently reproduce resolution proofs~\cite{PipatsrisawatD11} and CDCL (with random variable decisions) can efficiently find bounded-width resolution proofs~\cite{AtseriasFT11}.
The complexity-theoretic power of restart in SAT solvers was studied in~\cite{LiFVPG20}.

\subsection{Organization of this paper}

The rest of this paper is organized as follows. First, in \refSec{sec:preliminaries}, we introduce the notations of the field of SAT solving that we are using, give a short overview of the technique of conflict-driven clause learning, and give some statistical background, especially of survival analysis.
We proceed to describe the experimental setup in \refSec{sec:experiments}.
Finally, \refSec{sec:is-cdcl-useful} investigates whether clause learning is useful on average.
\refSec{sec:multimodal} demonstrates that the runtime distributions of CDCL solver configurations we investigated exhibit a multimodal behavior.
This investigation is continued in \refSec{sec:MultimodalImpliesMixedButWhichOne}, 
where it is shown that the Weibull mixture distribution is a suitable fit for the runtime distributions of these CDCL solvers.

\section{Preliminaries}
\label{sec:preliminaries}

A \introduceterm{literal} $\ell$ over a Boolean variable $x$ is either~$x$ itself or its negation $\overline{x} \coloneqq \neg x$.
A \introduceterm{clause} $C = (\ell_1 \lor \dots \lor \ell_k)$ is a (possibly empty) disjunction of literals $\ell_i$.
If a clause contains only one literal, it is called \introduceterm{unit}.
A \introduceterm{CNF formula} $\form = C_1 \land \dots \land C_m$ is a conjunction of clauses.
We also write clauses as a set of literals and CNF formulas as a set of clauses.
An \introduceterm{assignment}~$\alpha$ for a CNF formula~$\form$ is a function that maps some subset of the variables occurring in~$\form$ to $\{0,1\}$.
By naturally extending~$\alpha$ by the definition $\alpha(\overline{x}) \coloneqq \overline{\alpha(x)}$, we can define the result of applying~$\alpha$ to a clause~$C$, which we denote by~$C|_{\alpha}$: one deletes all occurrences of literals~$\ell$ from~$C$, where $\alpha(\ell)=0$; if there is a literal~$\ell \in C$ with $\alpha(\ell)=1$, then $C|_{\alpha} = 1$.
The notation~$\form|_{\alpha}$ denotes the formula where all clauses containing a literal~$\ell$ with~$\alpha(\ell) = 1$ are deleted, and each remaining clause~$C$ is replaced by~$C|_{\alpha}$.
A clause $C$ is called a \introduceterm{logical consequence} of a formula~$\form$ if, for all assignments~$\alpha$ with $\form|_{\alpha} = 1$, it also holds $C|_{\alpha} = 1$.
A set~$\Clauses$ of clauses is a logical consequence of~$\form$ if each clause $C \in \Clauses$ is a logical consequence of~$\form$. We then call the formulas $\form$ and $\form \cup \Clauses$ \introduceterm{logically equivalent}.

\subsection{Conflict-driven clause learning solvers}
\label{sec:cdcl}

\introduceterm{Conflict-driven clause learning} SAT algorithms, or \introduceterm{CDCL} for short, are one of the most remarkable success stories in computer science. Introduced in the works~\cite{MS99GRASP} and~\cite{MMZZM01Chaff}, CDCL can yield dramatic speedups over the simple recursive depth-first backtracking approach \algoformat{DPLL}~\cite{DP60AComputingProcedure,DLL62AMachineProgram}. The DPLL algorithm essentially selects an unassigned variable~$x$ of the formula~$\form$ it is trying to solve, and branches with calls to $\algoformat{DPLL}(F|_{[x = 0]})$ and $\algoformat{DPLL}(F|_{[x = 1]})$. 
While CDCL has been intensely studied by theoreticians and practitioners, we still do not have a complete understanding of \emph{all} mechanisms involved. %

\subsubsection{Clause learning}

In the following, we give a simple introduction to one of the most fundamental CDCL techniques: \introduceterm{clause learning}.
Informally speaking, this can be seen as a modification of~\algoformat{DPLL}, where the algorithm adds some clauses to~$\form$ if it reaches a conflict, \ie when the partial assignment constructed thus far falsifies a clause in~$\form$. The idea behind this is to prune the search tree and avoid having to re-explore some literal assignments that will not lead to a solution.

We introduce clause learning mostly by example, following the exposition in~\cite{ST13SATProblem}, and refer the reader to \cite{SLM21CDCL,BN21PCandSAT} for more details. As an example, consider as solver input the formula given in conjunctive normal form
    \begin{align}
        \begin{aligned}
        \label{eq:FormulaForCDCL}
            &(\overline{\varia} \lor \varib) \land
            (\overline{\varib} \lor \varic \lor \varis) \land
            (\overline{\varib} \lor \overline{\varit}) \land
            (\overline{\varis} \lor \varit \lor \variu) \, \land \\
            &(\overline{\variv} \lor \variw) \land
            (\overline{\variw} \lor \overline{\varix}) \land
            (\varix \lor \overline{\variy}) \land
            (\varic \lor \overline{\variw} \lor \variy).
        \end{aligned}
    \end{align}
Suppose that the CDCL solver makes its first \emph{decision} to assign $\varia = 1$.
The solver always looks out for clauses where all literals but one are assigned value 0 by the current assignment, and the remaining literal is unassigned (so-called \emph{unit clauses})~\cite{SLM21CDCL}
and assign this remaining literal so that the clause is satisfied.
This process is called \emph{unit propagation} and is repeated until there are no more unit clauses. In our example, using unit propagation, the solver sets $\varib = 1$ due to the clause $\left(\overline{\varia} \lor \varib\right)|_{[x_1 = 1]}$. It then sets $\varit = 0$ because of the clause $(\overline{\varib} \lor \overline{\varit})|_{[x_1 = 1,\:x_2 =1]}$. No more assignments can be made by unit propagation.
To move things further along, the solver has to make another decision. In our example, the solver now decides to set $\varic = 0$. By unit propagation, $\varis = 1$ and $\variu = 1$ are assigned.
Suppose, in its third decision, the solver sets $\variv = 1$. Using unit propagation, the assignments $\variw = 1$, $\varix = 0$, $\variy = 1$, \emph{and} $\variy = 0$ are made. This is a \emph{conflict} since the variable~$\variy$ cannot be set to both~$0$ and~$1$.

\begin{figure}[t]
    \centering
    \includegraphics[height=6cm]{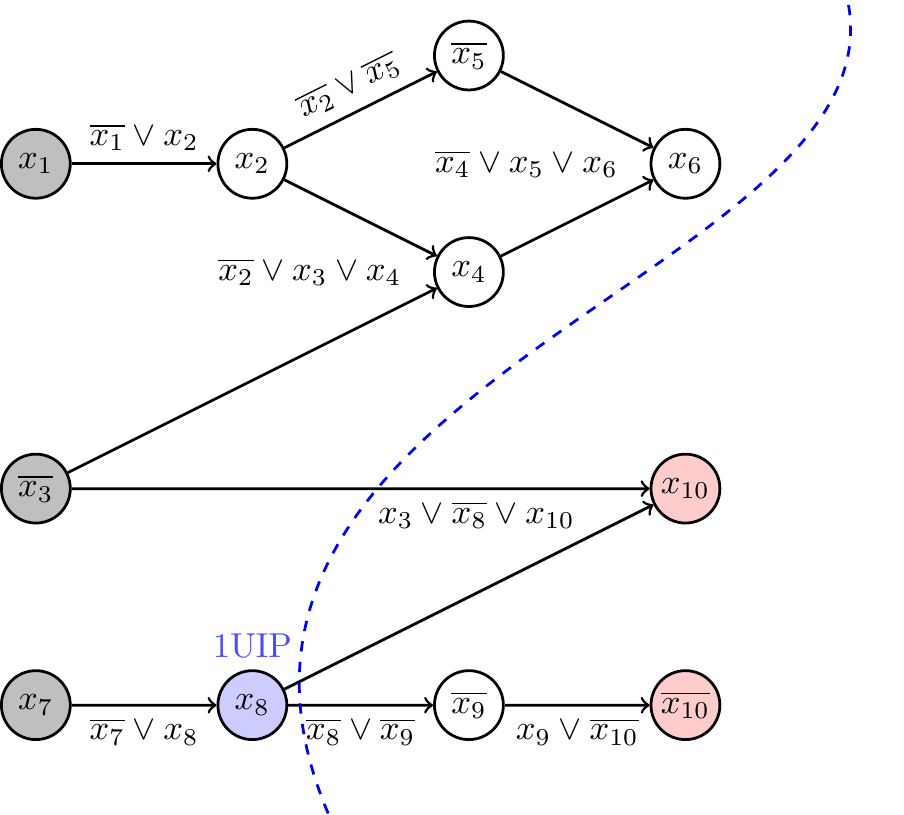}
    \caption{{\bf Conflict graph.}
    The figure shows the conflict graph generated in our example run of CDCL when solving formula~\eqref{eq:FormulaForCDCL}. The decision literals are marked in gray. The conflict literals are marked in red. The reason(s) for a propagation is given as label(s) of the edge(s). The first unique implication point (1UIP) is shown in blue. The 1UIP cut is the dashed blue line leading to the learned clause $(\varic \lor \overline{\variw})$. The graphic is adapted from~\cite{ST13SATProblem}.}
    \label{fig:ConflictGraph}
\end{figure}

During \emph{conflict analysis}, the solver learns a new clause. For this process, the solver uses the \emph{implication graph} that was built in stages during the execution of the algorithm (see \reffig{fig:ConflictGraph}). In this graph, \emph{decision literals} (in our example, $\varia$, $\overline{\varic}$, and $\variv$) are the source vertices. The \introduceterm{conflict literals} in our example are $\variy$ and $\overline{\variy}$. Furthermore, the graph includes vertices for every literal that has been assigned the value~$1$. A directed edge from node~$\vertexstd$ to node~$\vertexalt$ is included if the value of~$\vertexalt$ was set by unit propagation and $\overline{\vertexstd}$ occurs in the clause that was the \emph{reason} for variable~$\vertexalt$ being set.

The \introduceterm{level} of a decision variable $x$ is defined as the number of decision variables that have been assigned before~$x$ plus 1. %
Literals that are implied by unit propagation by a decision literal $x$ have the same level as $x$.
In our example, $\varia, \varib, \overline{\varit}$ have level~1; $\overline{\varic}, \varis, \variu$ are level~2; and $\variv, \variw, \overline{\varix}, \variy, \overline{\variy}$ are at the \introduceterm{conflict level}~3.

A vertex~$\vertexalt$ is called \introduceterm{UIP} (\introduceterm{unique implication point}), if all paths from the conflict level decision literal~$\variv$ to the conflict literals $\variy, \overline{\variy}$ run through~$\vertexalt$. Here, $\variv$ and $\variw$ are UIPs. The UIP closest to the conflict literals is called \introduceterm{first UIP} (\introduceterm{1UIP}) \cite{zhang2001efficient}. In our example, $\variw$ is the 1UIP.
While there are many ways to learn clauses from conflicts, the most popular method, invoked by a majority of modern CDCL solvers, is based on the 1UIP learning scheme.\footnote{This learning scheme is considered to work best (see~\cite{DershowitzHN07} for a comparative study of different learning schemes).} In this scheme,
the implication graph is uniquely cut such that
    \begin{enumerate}
        \item the 1UIP and all literals assigned before the conflict level are on one side,
        \item while all literals assigned after the 1UIP are on the other side (cf.~\cite{Mitchell05}).
    \end{enumerate}
This cut, shown in \reffig{fig:ConflictGraph}, yields the literals $\overline{\varic}$ and $\variw$ as starting points of the separated edges. The clause $\neg (\overline{\varic} \land \variw) = (\varic \lor \overline{\variw}) =: C$ can be shown to be a logical consequence of the original formula. This clause is called the \introduceterm{learned clause}. The solver adds this clause to the clause set. 

Using so-called \introduceterm{non-chronological backtracking}, the solver now jumps back to the level of the last variable in $C$ being assigned before the variable~$\variw$ of the conflict level, \ie it jumps back to level~2, where $\overline{\varic}$ was assigned. Then, using unit propagation on the clause~$C$, the variable~$\variw$ would be assigned~$0$.

\subsubsection{Clause deletion and literal block distance}

Many CDCL algorithms also incorporate a mechanism to delete (subsets of the) learned clauses from time to time.
As Mitchell explains in~\cite{Mitchell05}, this is due to the fact that even when sufficient memory is available, the time required to perform unit propagation becomes impractical for very large clause sets, thus reducing the performance of the solver.
However, deleting too many learned clauses breaks down the learning benefit~\cite{AS08OnTheGlucose}. To identify good learned clauses during the search, \cite{AudemardS09} introduced the notion of literal blocks. Intuitively, this measure tries to capture semantic relations between literals.

\begin{definition}[\cite{AudemardS09}]
    Let $C$ be a learned clause that is being falsified by the current partial assignment. Suppose $C$ is being partitioned into $k$ literal sets with respect to their level. The \emph{literal block distance} (\emph{LBD}) of the clause~$C$ is defined to be $k$.
\end{definition}

Clauses with higher LBD are considered to be less useful.
Other measures considered when deleting clauses are the size, age, and activity of a clause~\cite{DarwicheP21}.

\subsection{Statistical background} %
\label{sec:StatisticalBackground}

This section briefly introduces the statistical tools used in this paper.

\begin{definition}[\cite{norman1994continuous}]
	\label{def:cdf_quantile}
	Let $X$ be a real-valued random variable.
	\begin{itemize}
	\item
		Its \introduceterm{cumulative distribution function} (\introduceterm{cdf}) is the function $F \colon \R \to [0,1]$ with 
			\[
			F_X(t) \coloneqq \prob{X \leq t}. %
			\]
	\item
		Its \introduceterm{quantile function} $Q_X \colon (0,1) \to \R$ is given by
		    \[
		    Q_X(p) \coloneqq \inf \Setdescr{t \in \R}{F_X(t) \geq p}.
		    \]
	\item
	    If there is a non-negative, integrable function~$f_X$ such that 
	        \[
		    F_X(t) = \integralLow[u]{-\infty}{t}{f_X(u)},
		    \]
		then we call $f_X$ the \introduceterm{probability density function} (\introduceterm{pdf}) of~$X$.
	\item
	    The \introduceterm{survival function} of~$X$ is given by
	        \[
	        S_X(t) \coloneqq \prob{X > t} = 1 - F_X(t).
	        \]
	\end{itemize}
\end{definition}

We need the fact that the quantile function is the inverse of the cdf in the next section.

\subsubsection{Visual data analysis}
\label{sec:IntroToQQPlots}

To compare two probability distributions, we use the explorative graphical tool of \introduceterm{Q\nobreakdash--Q~plots}. These plots compare two distributions by plotting their quantiles against each other. If the result is a line, one can assume that the underlying distributions are the same.

\begin{definition}
    Let $F$ and $G$ be two cdfs. Then the graph $\big( F^{-1}(p), G^{-1}(p) \big)$ for $0<p<1$ is called \introduceterm{Q--Q plot} of $F$ and $G$.
\end{definition}

\begin{remark}[\cite{gibbons2014nonparametric}]
    If $F$ and $G$ are identical, the Q--Q plot is the main diagonal.
    If $F(x) = G \left( \frac{x-\mu}{\sigma} \right)$, then $F^{-1}(p) = \mu + \sigma G^{-1}(p)$. Thus, the Q--Q plot of $F$ and $G$ shows a linear relationship of slope~$\sigma$ and intersection~$\mu$.
\end{remark}

In a goodness-of-fit problem, one theoretical cdf is given, and we have empirical observations drawn from the other distribution.

\begin{definition}
    Given a sample $y_{(1)} \leq y_{(2)} \leq \dots \leq y_{(k)}$, we let $p_i \coloneqq \hat{F}_n(y_i)$ and $x_i \coloneqq Q(p_i)$, where $\hat{F}_n$ is the empirical cumulative distribution function and $Q$ is the theoretical quantile function of a theoretical distribution function~$F$. In the \introduceterm{Q--Q plot}, we plot the points $(x_i, y_i)$ for $i=1,\dots,k$.
\end{definition}

\subsubsection{Survival analysis and censored data}
\label{sec:SurvAna}

We use survival analysis (see \cite{aalen2008survival} for an introduction to the subject) to analyze data in which the time until an event is of interest. The time until this event happens is called \introduceterm{event time}.
If all events are observed, we can estimate the cdf with the help of the observations, for which we use the empirical cdf.

\begin{definition}
	Let $X_1, \dots, X_n$ be independent, identically distributed real-valued random variables
	with realizations $x_i$ of $X_i$.
	Then, the \introduceterm{empirical cumulative distribution function} (\introduceterm{ecdf}) of the sample $(x_1, \dots, x_n)$ is defined as
	\[
	\hat{F}_n(t) \coloneqq \frac{1}{n} \sum_{i=1}^n \Indikator{\set{x_i \leq t}}, \quad t \in \R,
	\]	
	where $\Indikator{A}$ is the indicator function of event $A$.
\end{definition}

Since in some of our experiments, it turned out to be computationally infeasible to wait until all formula instances are solved, we use a tool from non-parametric statistics to estimate the survival function of the corresponding runtime random variable. %
That is, we are working with incomplete observations.
To nevertheless estimate the survival function from a sample of censored survival data, we use the Kaplan--Meier product-limit estimator~\cite{bohmer1912theorie,kaplan1958nonparametric}.

Let $T$ be a non-negative random variable (which indicates the time until an event of interest takes place, \eg finding the solution of a formula).
Let $t_1, \dots, t_k$ be the points in time when events $1, \dots, k$ would have happened (think of a solution for formula~$\form_j$ being found if the solver was not stopped) whose common distribution is that of~$T$.
\introduceterm{Right-Censoring} is present when we have some information about event time (\eg the solver was still running at a certain point in time), but for some events, we do not know the exact event time (because we stopped the solver early).
More precisely, to avoid excessively long runtime, we later choose for every $j \in \set{1, \dots, k}$ a fixed integer $c_j$ as the \introduceterm{censoring time} for event $j$ (meaning that after this time, the solving of $\form_j$ is aborted).
Then, the data available for estimating the survival function~$S_T$ of the random variable~$T$ is the sequence of observations
    \[
        \big( (\widetilde{t}_j,c_j) \big)_{j=1,\dots,k}
        \quad \text{ with } \quad
        \widetilde{t}_j \coloneqq \min \set{t_j, c_j},
    \]
as well as \introduceterm{censoring indicators} $\mathrm{cen}_j \in \set{0,1}$ of the form
    \begin{align}
    \label{eq:Censoringj}
        \mathrm{cen}_j = 0 :\!\iff t_j < c_j.  
    \end{align}
That is, we either know that the formula $\form_j$ was solved in time (and we know the time~$t_j$ needed for this), or we know that the solver was still running at the censoring time~$c_j$.

\begin{definition}[\cite{bohmer1912theorie,kaplan1958nonparametric}]
\label{def:KaplanMeier}
    The \introduceterm{Kaplan--Meier estimator} is given by
        \[
        \widehat{S}_T(t)
        \coloneqq
        \prod_{i:\, t_i \leq t}
        \left( 1 - \frac{d_i}{n_i} \right),
        \]
    where (in our case)
        \begin{itemize}
            \item 
                $t_i$ is a point in time when (at least one) formula was solved,
            \item
                $d_i$ is the number of experiments, where the solver finished at time~$t_i$, and
            \item
                $n_i$ is the number of experiments that have not yet had an event or have not been censored up to time~$t_i$.
        \end{itemize}
\end{definition}

If there are no censored observations, the Kaplan--Meier estimator reduces to one minus the empirical cumulative distribution function (see \egcite~\cite{aalen2008survival}), also known as the empirical survival function.

\section{Experimental setup}
\label{sec:experiments}

There have been significant advances in the theoretical field of proof complexity developing a theoretical understanding of CDCL solvers, as we have surveyed in \refSec{sec:RelatedWork}.
Unfortunately, no model completely captures clause deletion in CDCL solvers.
For example, in \cite{BeameKS04, PipatsrisawatD11, AtseriasFT11}, the analyses of theoretical solvers rely crucially on the assumption that the learned clauses are never deleted.
For these reasons, an experimental approach seems the most reasonable to investigate the effect of pre-learned clauses and clause deletion.

Let us briefly summarize our experimental approach before discussing its details. First, we recorded all learned clauses~$\AllClauses$ a CDCL solver will find on ``its way towards'' a solution of a formula~$\form$; then, we extended the original instance with subsets $\Clauses \subseteq \AllClauses$ of these pre-learned clauses; and finally, we analyzed the runtime of such extended instances compared to the original instance~$\form$ (see \refsec{sec:gen-extensions} for additional details on and a discussion of this modification process).

We now describe our CDCL solver configurations and choice of benchmarks.

\paragraph{Investigated solvers and deletion strategies.}
Our SAT solver of choice for the majority of our experiments was \GlucoseFour{} (see~\cite{audemard2009glucose, AudemardS09}) in the 
non-parallelized version. 
First introduced in 2009, the \GlucoseProject{} project, which is based on the famous \MiniSAT{} solver~\cite{een2003extensible}, was quite successful in the past SAT Competitions. 
To increase confidence in the results, we also performed experiments using the deletion strategy of Chanseok~Oh in combination with the \GlucoseFour{} solver.
We furthermore extended the examination to the \MiniSAT{} solver.

\paragraph{Instances.}
We obtained a relevant, diverse, and well-documented pool of satisfiable and unsatisfiable instances by choosing all instances from the SAT Competition~2020, which were solved by \GlucoseThree{} between \SI{30}{\minute} and \SI{5000}{\second} ($\approx$~\SI{83}{\minute}).\footnote{\GlucoseThree{} was used for the filtering of the instances. We, however, later used \GlucoseFour{} for our experiments as this was the newest version of the solver that was available at the start of our experiments. This solver did not yet participate in the SAT Competition 2020.} The upper bound comes from a time limit imposed in the SAT Competition, where all solvers are cut off after 5000\,s. A more detailed description of all selected instances can be found in \nameref{S1_Table}. We also refer to the proceedings of the SAT Competition~2020~\cite{balyo2020proceedings, froleyks2021sat}.
A vigilant reader may notice that we have 53~instances in our pool, whereas our selection criterion applies to 61 instances of the SAT Competition 2020. We eliminated the remaining eight instances from the pool because they caused technical complications in at least one stage of our experimental setup. For example, three cases failed during clause recording as the number of learned clauses was too high and the required disk space to save all of them exceeded all reasonable limits. On the remaining five instances, \GlucoseFour{} ran out of RAM for some extensions. These cases could skew the runtime analysis since we do not know how \GlucoseFour{} would have performed with enough memory. Therefore, we excluded them from the analysis.

\subsection{Generating extensions from learned clauses}
\label{sec:gen-extensions}

During execution, modern CDCL solvers learn plenty of clauses.
All these learned clauses are directly implied by the clauses of the initial formula $\form$, which means that $\form \cup \Clauses$ is logically equivalent to~$\form$ for all $\Clauses \subseteq \AllClauses$, with $\AllClauses$ being the set of all learned clauses. We call $\Clauses$ an \introduceterm{extension} of the \introduceterm{base instance}~$\form$ and $\form \cup \Clauses$ an \introduceterm{extended instance}.

\begin{algorithm}[t]

    \SetKw{KwWithProb}{with probability}
    \SetKw{KwDo}{do}

	\textbf{Input:} Boolean formula~$\form$ (the base instance)
	\BlankLine
	Let $\AllClauses$ be the set of all learned clauses during the execution of \Algorithm$(\form)$\\
	$L \coloneqq \emptyset$\\
	\ForEach{$C \in \AllClauses$}{
	    \KwWithProb{$p$} \KwDo
	        $L \coloneqq L \cup \set{C}$\\
	    }
	Call \Algorithm$(\form \cup L)$ and record several performance measures %

	\caption{{\bf Modified version of a CDCL Solver.}
	We used this modified version of $\Algorithm{} \in \Set{ \GlucoseFour{}, \: \GlucoseFour{}\texttt{\,+\,ChanseokOh}, \: \MiniSAT{} }$ in our experiments to model the clause learning and clause deletion process as a random process.
	Each call of this modified algorithm uses \Algorithm{} to solve an extended instance~$\form \cup \Clauses$.
	This allows us to study the runtime distribution of \Algorithm{}.
	}
	\label{algo:mainNew}
\end{algorithm}

\paragraph{Our model.}

We adapt the approach presented in~\cite{WL21Evidence} to CDCL solvers. We refer to \refAlg{algo:mainNew}, which requires a pool of pre-learned clauses~$\AllClauses$. These clauses were gathered by running $\Algorithm{} \in \Set{ \GlucoseFour{}, \: \GlucoseFour{}\texttt{\,+\,ChanseokOh}, \: \MiniSAT{} }$ on a base instance~$\form$ and logging all learned clauses to a file.
Thus, the set~$\AllClauses$ contains all clauses \Algorithm{} learned ``on its way'' to a satisfying assignment or to an unsatisfiability proof.
The random sampling of a subset~$L \subseteq \AllClauses$ was implemented by independently selecting each clause in the pool~$\AllClauses$ with probability~$p$. 
This subset~$\Clauses$ is used to study the runtime of \Algorithm{} on the extended~instance~$\form \cup \Clauses$. 
For our experiments, we chose~$p = 0.01$ and generated $N$ different extensions $L^{(1)}, \dots, L^{(N)}$ for each of the 53~base~instances in our instance pool.
That is, for each base instance~$\form$, we recorded the performance measures (runtime, number of conflicts, number of propagations, etc.) of \Algorithm{} on the extended instances $\form \cup L^{(1)}, \dots, \form \cup L^{(N)}$.
We chose $N = 5000$ for \GlucoseFour{} and $N = 1000$ for the additional experiments with $\GlucoseFour{}\texttt{\,+\,ChanseokOh}$ and $\MiniSAT{}$. 
In this way, we can study the performance of \Algorithm{} on instances that \emph{already} contain some of the learned clauses.

The scripts for generating the sets~$\AllClauses$ corresponding to the base instances, as well as the scripts for reconstructing our sampled sets~$\Clauses$ can be found in \nameref{S1_File}.
Regarding only the experiments where $\Algorithm{} = \GlucoseFour{}$, we produced 1.5~TB of instance data in the DIMACS format.
Recording the performance, we used \num{265000} calls of \refAlg{algo:mainNew} with sometimes surprisingly long runtimes.

\paragraph{Interpretation of the model.}

Our model can be interpreted in the following two ways:
\begin{enumerate}[label=(\alph*)]
    \item
        We force the investigated solver \Algorithm{} to always keep the clauses in~$\Clauses$ in memory, while it is solving $\form \cup \Clauses$.
        This is interesting because the clauses in~$\Clauses$ were learned by the same solver on ``its way to'' a solution.
        This can be thought of as providing the solver these clauses for free by a \emph{useful} clause oracle.
    \item 
        The investigated solver \Algorithm{} starts by trying to solve the base instance~$\form$. During its execution, the solver learns the clauses of the core pool~$\AllClauses$. Let a \textbf{reset} of the solver be defined as an aggressive clause deletion of part of the learned clauses in the database with a simultaneous deletion of the current assignment. We model this reset as a probabilistic process that takes place just before a solution is found, \ie the solver deletes each of the learned clauses with probability $1-p$. The set~$\AllClauses$ could be thought of as being kept as a \emph{frozen set}. Our model studies the runtime distribution after this reset.
\end{enumerate}

One should observe that our sampling process was designed so that all extensions have comparable LBD properties.
For example, the median LBD of an extension is approximately the median LBD of the pool~$\AllClauses$.
The same holds for various LBD quantiles. 
Thus, long runtimes on extended instances cannot be explained by the fact that the set $L$ of the extended instance has ``poor'' LBD properties.
The same observation holds for the average size of the learned clauses in the extensions.
Long runtimes on extended instances are thus not due to the fact that these extensions have a higher number of clauses with more literals and are thus less likely to become unit and actually guide the solver.
Additionally, the sizes $| \form \cup \Clauses |$ follow a Gaussian normal distribution for fixed~$\form$. Even if a solver scales the size of the learned clause database with the size of the original problem and thus accumulates even more learned clauses, this observation is not enough to explain the multimodality and long-tailed phenomenons observed.

We furthermore would like to mention that we did not increase the base instance~$\form$ by too many clauses: the median increase of $|\form \cup \Clauses|$ when compared to $|\form|$ is just 3\,\%. This increase cannot explain the enormous deteriorations that we have observed.

\paragraph{Limitations of our model.}

One obvious criticism of the approach described above is that our model is not a 100\,\% precise modeling of CDCL solvers as they are implemented in practice.
For example, our experimental setup cannot preserve variable activities between the solving of $\form$ and $\form \cup L$. 
Our notion of a reset also assumes that the reset takes place just before a solution is found. In practice, a deletion of learned clauses is usually triggered much earlier.
However, to conduct a thorough statistical analysis of runtime distributions, we had to make sure that the set~$\AllClauses$ is large enough to sample at least 1000 quite different extensions. Our approach also has the benefit of studying the combined influence of learned clauses from far apart \emph{batches} (a batch is the set of clauses learned between restarts\footnote{A restart of a CDCL solver consists of deleting the current partial assignment while keeping the set of learned clauses. This concept should not be confused with our notion of a reset.} of the solver). We believe that our approach lends itself to getting a more \emph{global} understanding of the role of learned clauses. 

While these are valid points, we nevertheless hope that our runtime distribution study provides new insights into the role of clause deletion.
The aim of our model is not to completely understand all aspects of CDCL solvers but to get a better understanding of one of them.
At the current state of research on CDCL solvers, clearly, all such models will have their drawbacks and will not be able to represent all heuristics correctly.
Further points that could be investigated or improved upon in future research are noted in \refsec{sec:Conclusion}.

\subsection{The challenge of solving a myriad of hard formulas}

Solving over \num{265000} hard Boolean formulas in a reasonable time required parallelization, for which we used \texttt{Sputnik}~\cite{VolkelLSKK15}, and a somewhat more complex experimental setup. We additionally distributed the formulas over two regular servers (\texttt{Luna}, \texttt{Erpel}) and an HPC cluster (\texttt{BwUniCluster}~\texttt{2.0}). See \reftab{tbl:hardware} for more details.\footnote{Note that due to such a diverse hardware setup, runtime comparisons have to be done with caution. More details on this and other metrics can be found in Section~\ref{sec:is-cdcl-useful}. During our experiments, we always additionally considered time-independent theoretical measures.}

After we initially started the experiments on just the \texttt{Luna} Server, we were confronted with surprisingly long runtimes on certain extended formulas. The corresponding original instances were solved after at most 95 minutes, but for some extended formulas, it took \GlucoseFour{} more than ten days to solve them. This led us to
    \begin{enumerate}[label=(\alph*)]
        \item\label{list:expr1} distribute the calls of Algorithm~\ref{algo:mainNew} over multiple hardware nodes, and
        \item\label{list:expr2} introduce a timeout strategy (censoring) as introduced in \refSec{sec:SurvAna}.
    \end{enumerate}
Only 12 out of all 53 instances encountered censoring in the \GlucoseFour{} experiments, meaning that the solver reached the timeout limit for at least one extended instance. In most cases, the solving of only a few extended instances had to be stopped due to the timeout policy. More details on the number of censored extended instances can be found in \nameref{S1_Table}.
Said censoring timeouts $c_j$ were assigned to each extended formula~$\form_j$ according to an offset geometric distribution such that
    \begin{align*}
        c_j \sim \Geo{\frac{1}{\SI{12}{\hour}}} + \SI{5000}{\second}.  
    \end{align*}
We have chosen the value of \num{5000}~seconds since this is the cut-off in the SAT Competition. The distribution $\Geo{\frac{1}{\SI{12}{\hour}}}$ has an expectation of \num{12} hours. 
This approach led to a total CPU time of \num{16}~years and \num{8}~month.
All raw data obtained in this way can be found in \nameref{S1_File}.

Having used censoring in acquiring our data makes the use of survival analysis, as elaborated in \refSec{sec:SurvAna}, necessary.

{
\renewcommand{\arraystretch}{1.25}
\begin{table}[t]
\centering
\caption{
    {\bf Summary of the hardware used in our experiments.}
}
\begin{tabularx}{\textwidth}{l l X r r r}
  \toprule
  Name & Node & CPU & Cores & Frequency & RAM\\
  \midrule
  \multicolumn{2}{l}{\texttt{Erpel}} & Intel Xeon E5-2698 v3 & 32 & 2.30 GHz & 256 GB \\
  \multicolumn{2}{l}{\texttt{Luna}} & AMD EPYC 7742 & 64 & 2.25 GHz & 256 GB \\
  \multirow{2}{*}{\texttt{BwUniCluster}~\texttt{2.0}} & HPC & Intel Xeon Gold 6230 & 40 & 2.10 GHz & 96 GB \\
                                    & HPC Broadwell & Intel Xeon E5-2660 v4  & 28 & 2.00 GHz & 128 GB \\
  \bottomrule
\end{tabularx}
\begin{flushleft}
\small{
    Table notes. \texttt{Erpel} and \texttt{Luna} are standard server architectures, whereas the \texttt{BwUniCluster}~\texttt{2.0} is a high-performance computing cluster.
    }
\end{flushleft}
\label{tbl:hardware} 
\end{table}
}

\section{Is clause learning useful on average?}
\label{sec:is-cdcl-useful}

Clearly, augmenting a basic DPLL solver with a clause learning mechanism is far from a modern CDCL solver. However, clause learning is arguably the most important technique in CDCL solvers, lending its name to the paradigm.
One would therefore expect that clause learning (especially when guided by state-of-the-art heuristics) is generally useful, \ie one would expect that providing the solver with learned clauses for free does increase the performance of the solver when compared to the base instance, where the solver has to learn all clauses by itself (we use interpretation~(a) of our model in this section).

To check this assumption, we performed a parametric test of whether the mean difference between the base instance and the 5000 (possibly censored) runtimes on the extended instances solved with \GlucoseFour{} equals 0.
For this, we assumed that the paired differences follow a Gaussian normal distribution.
To perform this test, we used the ``NADA2: Data Analysis for Censored Environmental Data'' package~\cite{NADA2} in \texttt{R} (we refer to the book~\cite{helsel2012statistics} for an in-depth treatment of the statistical methods involved).
For the threshold value in the statistical null-hypothesis tests, we used the standard value $p=0.05$.

In this section, we---quite surprisingly---demonstrate that clause learning is oftentimes useful, but there are also \emph{many} instances where a dramatic negative effect can be observed.
We speak of a \introduceterm{negative effect} if the mean of the runtimes required to solve the extended instances was statistically significantly greater ($p=0.05$) than the runtime required to solve the base instance.
We refer to \reftab{tbl:tTestZeit} for an overview of the different effects.
Interestingly, almost all instances can be very clearly categorized in the table because the obtained $p$-values are remarkably low (the exceptions are marked in the table).

\begin{table}[t]
\centering
\caption{
    {\bf The effect of learned clauses 
    (without deletion) 
    on the runtime of \GlucoseFour{}.}
}
\begin{tabular}{lcc}
  \toprule
                        & With censoring            & Without censoring\\
  \midrule
  Positive effect       & 5 & 26 \phantom{[$\ddagger$]} \\
  No significant effect & 0     & \phantom{1}1 [$\dagger$]  \\
  Negative effect       & 7             & 14 [$\ddagger$]            \\
  \bottomrule
\end{tabular}
\begin{flushleft}
\small{
    Table notes.
    We say that clause learning has a \emph{positive effect} if the mean of the runtimes required to solve the extended instances was statistically significantly smaller ($p = 0.05$) than the runtime required to solve the base instance. 
    If the mean is statistically significantly greater than the runtime for the base instance, we speak of a \emph{negative effect}.
    Otherwise, it has \emph{no significant effect}.
    The base instances are grouped in the table based on whether there was at least one extended instance of this base instance where censoring occurred (\emph{with censoring}) or not (\emph{without censoring}).
    $\vartriangleright$ [$\dagger$] The observed $p$-value was $p = 0.53$ for the unsatisfiable instance $\instanceformat{ncc\_none\_5047\_6\_3\_3\_0\_0\_41\_p0.01}$ (see \nameref{S1_Table} for the complete list of $p$-values). The observed effect was negative.
    $\vartriangleright$ [$\ddagger$] Furthermore, the non-censored instances with negative effects include an instance (namely \instanceformat{6g\_5color\_164\_100\_01}) with $p = 0.039$. This $p$-value is noteworthy since all other $p$-values are smaller than~$2.657 \cdot 10^{-4}$ (again, see \nameref{S1_Table}).
    }
\end{flushleft}
\label{tbl:tTestZeit}
\end{table}

It is worth mentioning that in~\cite{ChanseokOh15}, a fundamental difference in the way modern CDCL solvers solve satisfiable and unsatisfiable instances was reported. 
There is still no consensus in the community regarding the role learned clauses play in obtaining a satisfying assignment. However, most agree that the role of learned clauses is more pronounced when the solver is constructing a proof of unsatisfiability. By analyzing the effect of learned clauses with respect to the satisfiability of the formula, we found that in 63\,\% of the cases, the effect was negative for satisfiable instances and in 27\,\% for unsatisfiable instances (see \reftab{tbl:tTestSAT}).

\begin{table}[t]
\centering
\caption{
    {\bf The effect of learned clauses 
    (without deletion)
    on the runtime of \GlucoseFour{} with respect to the satisfiability of the base instance.}
}
\begin{tabular}{lcc}
  \toprule
                        & Satisfiable            & Unsatisfiable \\
  \midrule
  Positive effect       & \phantom{1}7 \phantom{[$\ddagger$]} & 24 \phantom{[$\dagger$]} \\
  No significant effect & \phantom{1}0 \phantom{[$\ddagger$]}     & \phantom{2}1 [$\dagger$]  \\
  Negative effect       & 12 [$\ddagger$]             & \phantom{2}9 \phantom{[$\dagger$]}            \\
  \bottomrule
\end{tabular}
\begin{flushleft}
\small{
    Table notes.
    See \reftab{tbl:tTestZeit} for an explanation of the rows.
    The dagger-symbols have the same meaning as in \reftab{tbl:tTestZeit}.}
\end{flushleft}
\label{tbl:tTestSAT}
\end{table}

We want to emphasize that there are also quite a few instances where adding the set~$L$ yields a deterioration in the number of conflicts occurring during a run of the solver.
In more than 11\% of the cases, an adverse effect was observed, which cannot be explained by pure chance and is quite surprising.
We refer to \reftab{tbl:tTestKonflikte} for an analysis of the effect of learned clauses concerning the number of conflicts.
We chose to study this additional measure due to the heterogeneous server architecture outlined in \reftab{tbl:hardware}: The runtimes differ very slightly between the server; however, the number of conflicts needed to solve an instance is a robust, hardware-independent measure.

\begin{table}[t]
\centering
\caption{
    {\bf The effect of learned clauses 
    (without deletion) 
    on the number of conflicts used by \GlucoseFour{}.}
}
\begin{tabular}{lcc}
  \toprule
                        & With censoring            & Without censoring\\
  \midrule
  Positive effect       & 8  & 37 \\
  No significant effect & 0    & \phantom{3}0 \\
  Negative effect       & 2            & \phantom{3}4            \\
  \bottomrule
\end{tabular}
\begin{flushleft}
\small{
    Table notes. 
    See \reftab{tbl:tTestZeit} for an explanation of the rows and columns (replace \emph{runtime} by \emph{number of conflicts}).
    $\vartriangleright$ Out of the six instances with a negative effect, two were unsatisfiable, and four were satisfiable.
    $\vartriangleright$ Two of the 53 instances (both with censoring) had to be excluded from the tests due to numerical complications in the censored data paired $t$-test (NADA2 package).
    }
\end{flushleft}
\label{tbl:tTestKonflikte}
\end{table}

For a comparison of runtime vs.\ number of conflicts, we refer to \reftab{tbl:tTestSignMultiplikation}.
We want to point out that in each case where an opposite effect for time and number of conflicts can be observed, the effect was negative for time and positive for the number of conflicts.
As mentioned in \refSec{sec:RelatedWork}, this might be explainable by the observation of Mitchell~\cite{Mitchell05}: the required time to perform unit propagation becomes too high for very large clause sets, which reduces the performance of the solver. 
If an extended instance already contains more clauses than the base instance, the propagation speed when solving the extended instance is lower. Furthermore, many solvers scale the size of the learned clause database with the size of the original problem, leading to the solver to learn even more clauses, which causes the propagation speed to lower again.

\begin{table}[t]
\centering
\caption{
    {\bf Comparison between the effect of adding clauses (without deletion) for time and number of conflicts of \GlucoseFour{}.}
}
\begin{tabular}{lcc}
  \toprule
                        & With censoring            & Without censoring\\
  \midrule
  Same effect for time and number of conflicts       & 7  & 30 \\
  Opposite effect & 3    & 10 \\
  \bottomrule
\end{tabular}
\begin{flushleft}
\small{
    Table notes. 
    We excluded one instance where, for at least one measure, our experiments could not determine if there was a positive or negative effect for that measure.
    }
\end{flushleft}
\label{tbl:tTestSignMultiplikation}
\end{table}

In both cases, \ie regardless of whether one considers the runtime or the number of conflicts, the observed performance deterioration cannot be explained by pure chance. 
This is all the more surprising since the set of learned clauses~$L$ is not just any random set of clauses but clauses that the same solver learned on ``its way to'' a solution. These clauses should thus contain very useful information for the solver when it is solving the same formula from scratch but with this additional information. Therefore, one would expect that each and any of such a clause would benefit the guidance of~CDCL towards a solution.

To explain this deterioration phenomenon, one should therefore consider the influence of clause \emph{deletion}. Our experimental setup can be interpreted as switching off clause deletion for the set of added clauses~$L$ (while keeping all other heuristics and optimizations of the solver) and learning all those clauses at once. Note that the solver learns some additional clauses during its run and can delete them. The set~$L$, however, is fixed during the run. Seemingly, clause deletion at the right points in time is as crucial as clause learning. This statement cannot be fully explained by a blow-up in the size of the clause database, as the unit-propagation-independent measure of the number of conflicts also increased in many cases.

\section{Multimodal behaviors in CDCL solvers}
\label{sec:multimodal}

In Section~\ref{sec:is-cdcl-useful}, our focus was to compare the behavior of \GlucoseFour{} on the unmodified base instance to the behavior on the modified instances that extended this base instance. From now on, we focus on studying the distributions of the modified instances with respect to different measures (see interpretation (b) of our model).

Our precise aim in this section is to get an understanding of the modality of the ensuing CPU time distributions. Recall that in statistics, a probability distribution with a single peak is called \introduceterm{unimodal}; otherwise, we speak of a \introduceterm{multimodal} distribution (a more formal definition can be found in~\cite{hartigan1985dip}).
An easy way to inspect the modality of a distribution is to inspect the histogram of the distribution visually. This method has the additional advantage that no statistical test has to be used that can distinguish between unimodal and bimodal distributions but rely on the knowledge of the underlying distribution type (\ie one does not need to know in advance if the distribution can be resolved into, \eg \emph{normal} distributions~\cite{pearson1894}).

Since our obtained data points are censored, we cannot immediately plot the histogram. To overcome this obstacle, we have used the Kaplan--Meier estimator (see Definition~\ref{def:KaplanMeier}) implemented in the Survival package~\cite{survivalpackage} in \texttt{R} to obtain a fit of the underlying survival function.
The Kaplan--Meier fit can be computed without prior knowledge of the underlying distribution.
Graphically, the Kaplan--Meier survival curve is a step function with a drop each time the solver has finished an instance.
The points where a drop can be observed can thus be used as an estimation basis to create the histogram. Note that in the improbable event that two instances take precisely the same time to be solved (while solving times are often very large), the resulting histogram underestimates the number of instances in the corresponding bin. However, this kind of event occurs so seldom that, for all intents and purposes, we can be satisfied with the obtained estimation of the actual histogram.

For a more detailed investigation of multimodal behavior, we have used Hartigans' dip test statistic~\cite{hartigan1985dip}. This test determines the amount of multimodality by ``the maximum difference, over all sample points, between the empirical distribution function, and the unimodal distribution function that minimizes that maximum difference''.\footnote{More formally, let $\rho(F, \mathscr{A}) := \inf_{G \in \mathscr{A}} \| F - G \|_{\infty}$ for any function space $\mathscr{A}$ of bounded functions. The \introduceterm{dip} of a cdf $F$ is given by $D(F) := \rho(F, \mathscr{U})$, where $\mathscr{U}$ is the class of unimodal distribution functions~\cite{hartigan1985dip}.} Phrased differently, the dip of a distributions function measures departure from unimodality. 
We have used a threshold dip value of $0.005$ to classify multimodal histograms: All histograms with a dip value over $0.005$ can clearly be seen to exhibit multimodal behavior, using only visual inspection. We want to emphasize that this cut-off value is rather strict (see \eg \reffig{fig:MultiPropDec}\,(b) which is clearly multimodal but below the threshold value).

\begin{figure}[t]
    \centering
    \includegraphics[width=\textwidth]{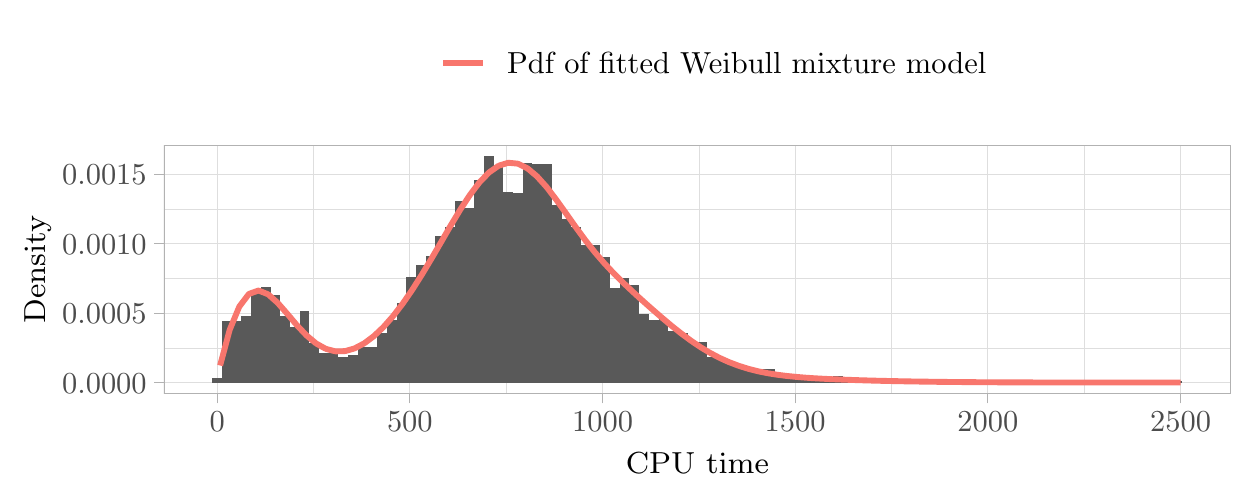}
    \caption{{\bf Multimodal histogram of runtime distribution.}
    We used the Kaplan--Meier estimate to obtain the histogram of the runtime distribution of the instance \instanceformat{UNSAT\_ME\_seq-sat\_Thoughtful\_p11\_6\_59-typed.pddl\_43}. We used the \introduceterm{Expectation--maximization} (EM) method to obtain the pdf of the fitted Weibull mixture model (see Definitions~\ref{def:Mixed} and~\ref{def:weibull} for an introduction to this kind of distribution). The EM algorithm is an algorithm that allows cluster analysis by starting with a heuristically initialized model and alternating between two steps. First, in the expectation-step (\introduceterm{E-step}), the association of the data points to the different clusters gets changed. Then, in the maximization-step (\introduceterm{M-step}), the model's parameters get improved by using this new association of the data points. We refer to the classic paper~\cite{dempster1977maximum} for an introduction to the algorithm. 
    The resulting fitted distribution that is seen in the plot is clearly multimodal. This is supported by a Hartigans' dip test value of $0.015 > 0.005$.}
    \label{fig:Histogram}
\end{figure}

\begin{figure}[t]
    \centering
    \includegraphics[width=\textwidth]{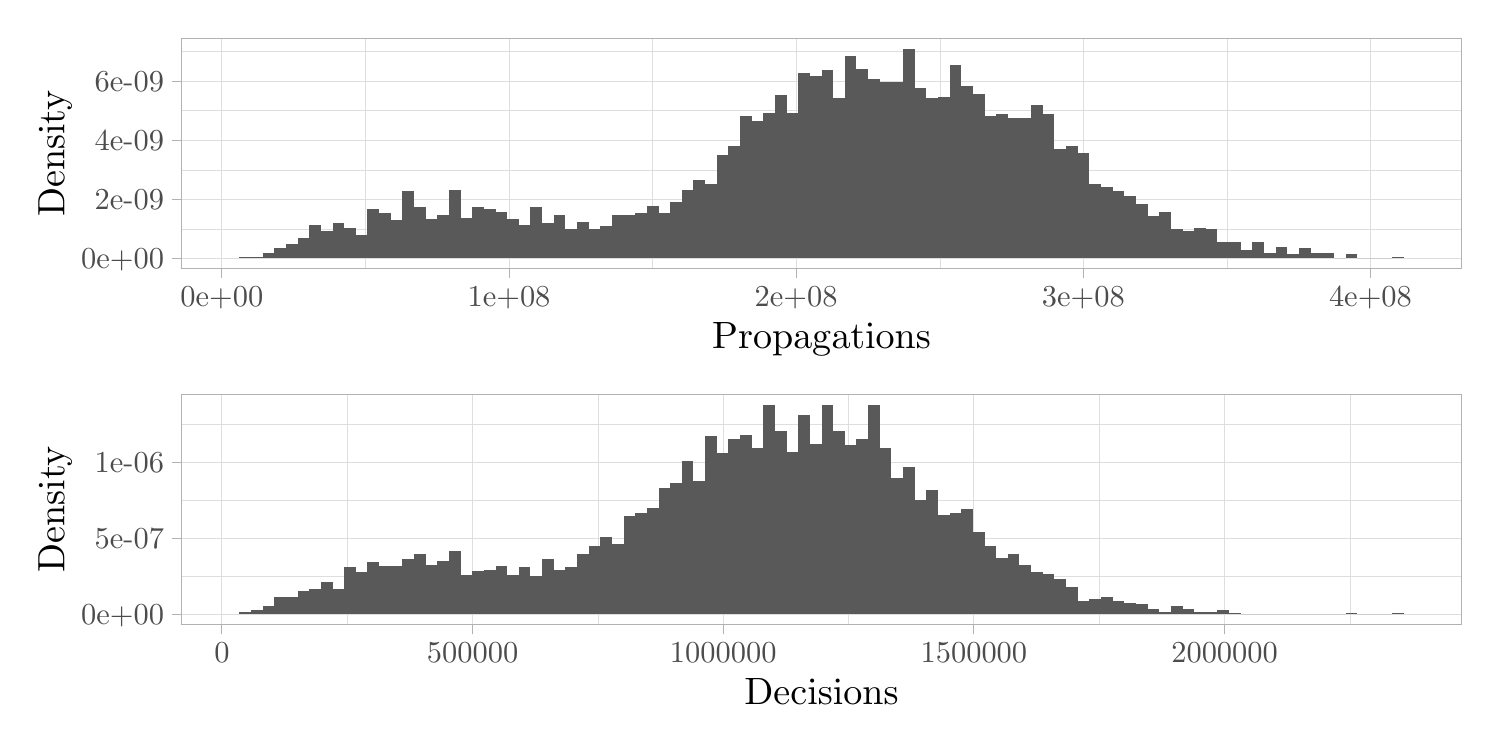}
    \caption{{\bf Multimodal histogram of the distribution for the number of propagations and decisions.}
    The histograms show the distribution of the two measures propagations and decisions required to solve the extended instances of \instanceformat{UNSAT\_ME\_seq-sat\_Thoughtful\_p11\_6\_59-typed.pddl\_43}. Both histograms for the hardware-independent measures have the same multimodal form as the histogram for CPU time shown in \reffig{fig:Histogram}.
    \textbf{(above)} Histogram for number of propagations with a Hartigans' dip test value of 0.005.
    \textbf{(below)} Histogram for number of decisions with a Hartigans' dip test value of 0.004.}
    \label{fig:MultiPropDec}
\end{figure}

We have printed the resulting estimated histogram using Kaplan--Meier of a representative instance in \reffig{fig:Histogram}. As can be clearly seen, the distribution is multimodal, as witnessed by the two peaks. The Hartigans' dip test value also confirms this.

\begin{figure}[t]
    \centering
    \includegraphics[width=\textwidth]{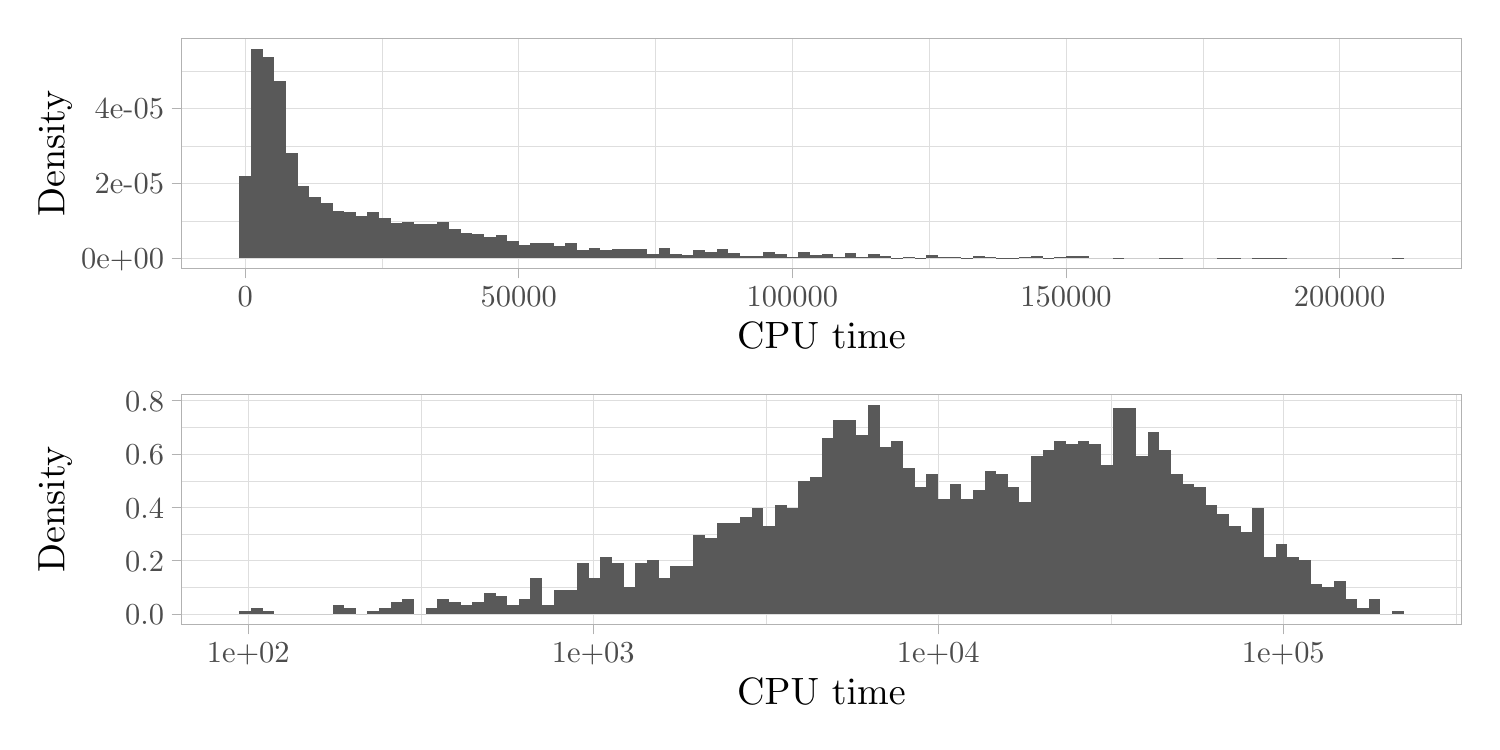}
  
    \caption{{\bf Histogram of runtimes vs.\ histogram of logarithmically scaled runtimes.}
    Scaling the $x$-axis of a histogram logarithmically can often uncover multimodality that is not clearly visible in the unscaled histogram. The graphic depicts both histograms for the instance \instanceformat{size\_5\_5\_5\_i019\_r12}, where this difference is very pronounced.
    \textbf{(above)} Histogram of CPU times with a Hartigans' dip test value of 0.005.
    \textbf{(below)} Histogram of logarithmically scaled CPU times with a Hartigans' dip test value of 0.016.
    }
    \label{fig:MultimodHistLog}
\end{figure}

To facilitate our inspection of the histograms, we also investigated the histograms for the logarithmically scaled runtimes. This method has been found to usually give a clear separation into a visible multimodal histogram if the underlying distribution is indeed multimodal (see \egcite~\cite{sedimentmultimodal1,sediment2} for the earliest applications of this technique).
In \reffig{fig:MultimodHistLog} we have shown an instance where the multimodality cannot be seen in the histogram at first glance but can clearly be made out in the histogram with logarithmically scaled runtimes.
Using this technique, we found that a significant fraction of 25\,\% of the instances exhibited dominant multimodal behavior.

Due to the heterogeneous server architecture, we also studied the histograms for hardware-independent measures, like the number of propagations and decisions needed to come to a solution.
Again, we could observe the multimodality phenomenon for these measures.
We refer to \reffig{fig:MultiPropDec} as a continuation of the histogram shown in \reffig{fig:Histogram}.

Using the threshold value of 0.005 for Hartigans' dip test, 32\,\% of all instances exhibit multimodal behavior for at least one measure (CPU time, logarithmized CPU time, propagations, or decisions).

This multimodal grouping of instances into several categories could be helpful in an investigation of the usefulness of the added clauses.
We make this thought more precise in the next section and investigate the distribution type underlying the model.

\section{Finding the right mixture distribution type}
\label{sec:MultimodalImpliesMixedButWhichOne}

In \refSec{sec:multimodal}, we have already seen that the runtime behavior is multimodal for a substantial part of the instances. This section aims to study which types of distributions are suitable to describe this behavior. Since most well-known distributions, such as the normal distribution, are unimodal or at most bimodal, this suggests that one must resort to another type of distribution.

The presence of the many peaks indicates that the different extended instances (and underlying clauses) can be divided into categories. Each category corresponds to the hardness of the extended instance, where the hardness is again not a fixed value but a random variable. 
If one confines oneself to a single category of extended instances, one is (potentially) no longer confronted with multimodal behavior but can describe the remaining data utilizing a unimodal distribution.

By (randomly) adding the clauses to~$L$, we then end up in this category of extended instances with a certain probability. If such an analysis is conducted for each category, we eventually obtain a description of the complete runtime behavior. Specifically, this means that for each category, the underlying runtime distribution, as well as the probability of ending up in that category, must be identified. The runtime behavior across all extended instances is then characterized by a so-called finite mixture distribution.

\begin{definition}
\label{def:Mixed}
    Let $X$ be a random variable having the cdf~$F_X$.
    Let $F_1, F_2, \dots, F_{\numberofcomponents}$ be cdfs 
    and $p_1, p_2, \dots, p_{\numberofcomponents}$ be weights 
    with $p_i > 0$ for all~$i \in \set{1, \dots, \numberofcomponents}$ 
    and $\sum_{i=1}^{\numberofcomponents} p_i =1$. 
    If
        \begin{align*}
            F_X(x) = \sum_{i=1}^{\numberofcomponents} p_i \cdot F_i(x)
        \end{align*}
    holds for all $x \in \R$, then $X$ has an \introduceterm{$\numberofcomponents$-component (finite) mixture distribution}.
\end{definition}

\begin{figure}[t]
    \centering
    \includegraphics[width=\textwidth]{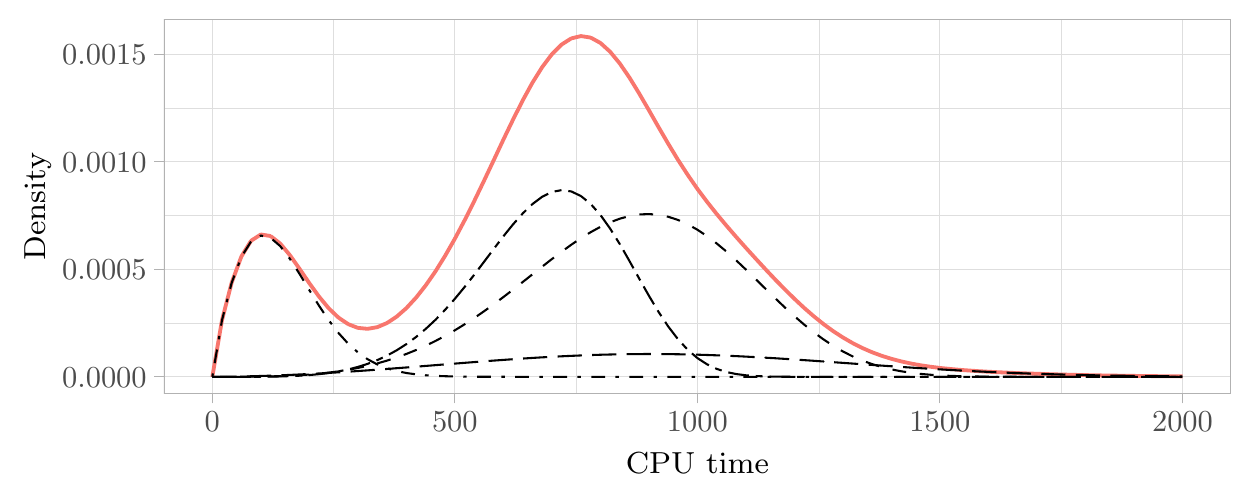}
    \caption{{\bf Mixture distribution and components.}
    The figure shows the Weibull components (in black) underlying the mixture distribution (in red) of instance \instanceformat{UNSAT\_ME\_seq-sat\_Thoughtful\_p11\_6\_59-typed.pddl\_43}. The Weibull components were scaled according to their respective $p_i$-values. For the histogram of this instance, refer to \refFig{fig:Histogram}.}
    \label{fig:Mixture}
\end{figure}

In our case, $\numberofcomponents$ can be understood as the number of categories. Furthermore, $p_i$ describes the probability of ending up in category~$i$, in which case $F_i$ is the runtime distribution for category~$i$. 
We refer to \refFig{fig:Mixture} of a depiction of the components underlying a mixture distribution.

In principle, one can choose arbitrary cdfs $F_1$ to $F_{\numberofcomponents}$. However, it is common practice to choose the cdfs from the same family of (parametric) distributions, where only the parameters of the distribution differ. For example, a popular model are Gaussian mixture distributions in which the cdfs describe normal distributions varying with respect to their expected value and variance. Therefore, the question arises of which parametric distribution type should be chosen for the mixture distribution. We shall argue that a distribution based on the so-called Weibull distribution is an appropriate type of distribution.

\begin{definition}[\cite{rinne2008weibull}]
	\label{def:weibull}
	A random variable $X$ with a pdf given by
    	\begin{align*}
        	f_X(x)=\begin{cases}
        	\frac{k}{a}\cdot \left(\frac{x-\ell}{a}\right)^{k-1}\cdot \e^{-\left(\frac{x-\ell}{a}\right)^k},& x \geq 0\\
        	0,& x < 0
        	\end{cases}
    	\end{align*}
	is \introduceterm{3-parameter Weibull distributed} with parameters $k \in \Rpos$ (\introduceterm{shape}), $a \in \Rpos$ (\introduceterm{scale}), and $\ell \in \R$ (\introduceterm{location}). The cdf of~$X$ is given by
    	\begin{align*}
        	F_X(x)=
        	\begin{cases}
        	1-\e^{-\left(\frac{x-\ell}{a}\right)^k},& x \geq 0\\
        	0,& x < 0.
        	\end{cases}
    	\end{align*}
	If the location parameter~$\ell$ is zero, we call the distribution \introduceterm{2-parameter Weibull distributed} or just \introduceterm{Weibull distributed}. 
\end{definition}

We first start by analyzing instances that can be described with only one component (\ie a 1-component mixture distribution). The idea behind this is that one can derive information about the instances that require more than one component. A suitable graphical tool for this analysis is provided by Q--Q plots, where the observed quantiles are plotted against the theoretical quantiles of a given distribution (recall \refSec{sec:IntroToQQPlots}). In the following, we consider the required CPU time until the respective instance is solved, \ie either a satisfying assignment is constructed, or a proof of unsatisfiability is established. 

\begin{figure}[t]
    \centering
    \includegraphics[width=0.5\textwidth]{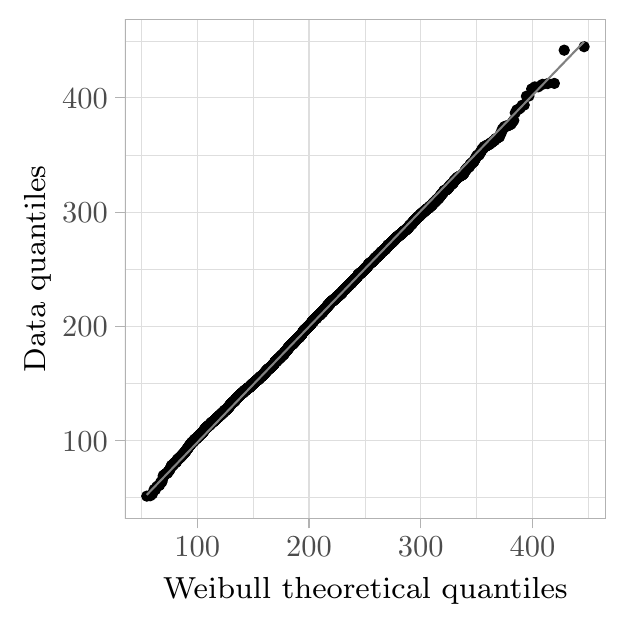}
    \caption{{\bf Q--Q plot.}
    The Q--Q plot for instance \instanceformat{crafted\_n11\_d6\_c4\_num19} was obtained by the quantiles of a fitted 3-parameter Weibull distribution and the data quantiles. The plot appears as a straight line. The correlation coefficient calculates to 0.9997979. For reference, the identity is given in gray.}
    \label{fig:qqwb3}
\end{figure}

As an example, we consider the data from one instance in \refFig{fig:qqwb3}. We use a fitted 3-parameter Weibull distribution as the theoretical distribution. As can be seen in the figure, the Q--Q plot yields a straight line, indicating that the theoretical distribution can describe the empirical data very well. So we can conclude that a 3-parameter Weibull distribution is a suitable description for this instance. This is not only the case in this example. For example, one can use the correlation coefficient to measure how linear a certain relationship is. A correlation coefficient of 0.999 describes an extremely strong linear relationship. We used this value and examined all Q--Q plots. %
In total, 12 instances reach a correlation coefficient of~at~least~0.999 if a fitted 3-parameter Weibull distribution is used as theoretical distribution. This suggests that a substantial number of instances can be described by a \emph{single} Weibull distribution. It should also be emphasized that other typical distribution types, such as the normal or lognormal distribution, do not provide good fits. While Weibull distributions describe a considerable fraction of the instances, this begs the question of what to do with the remaining instances.

First, it should be emphasized that especially the instances where multimodality is strong cannot be described by a single Weibull distribution. Then, however, it is natural to assume that the individual components of a mixture distribution follow Weibull distributions. We pursue this line of thought in more detail in the following.

Graphical analyses are well suited to argue that Weibull distributions are appropriate. First, we examine how the Weibull distribution behaves at the left tail, \ie the behavior if $x$ approaches $0$. It is well known that Weibull distributions have a linear appearance on a $\log$--$\log$ plot of the cdf at the left tail. To see this, we investigate the logarithm of the Weibull cdf~$F$ with location~parameter~$\ell = 0$ for $x \geq 0$:
    \begin{align*}
        \log F(x) = \log \left( 1-\e^{-\left(\frac{x}{a}\right)^k} \right).
    \end{align*}
Plugging the Taylor expansion of $\exp(-x)$ into this equation yields:
    \begin{align*}
        \log F(x) &= \log \left( 1- \left[ 1 -\left(\frac{x}{a}\right)^k + \frac{(\frac{x}{a})^{2k}}{2!} - \dots \right]    \right)\\
        &= \log \left( \left(\frac{x}{a}\right)^k - \frac{\left(\frac{x}{a}\right)^{2k}}{2!} + \dots     \right).
    \end{align*}
Considering the behavior as $x$ approaches~$0$, we notice that the trailing terms approach~$0$ much faster than~$(x/a)^k$ and thus can be neglected.
Hence, we obtain:
    \begin{align*}
        \log F(x) \approx  \log \left( \left(\frac{x}{a}\right)^k \right) = k \cdot \log x - k\cdot \log a \quad \text{for } x \to 0.
    \end{align*}
By substituting $z=\log x$, one finds that the cdf~$F$ indeed appears linearly in the neighborhood of zero on a $\log$--$\log$ plot.

For mixture distributions, this method is useful for making statements about the smallest component. Suppose that $F(x) = \sum_i^{\numberofcomponents} p_i F_i(x)$ is the cdf of a mixture distribution. Here, $F_1$ is the cdf of a Weibull distribution, and for small $x$, we have $F_1(x) \gg 0$ and $F_2(x) \approx F_3(x) \approx \dots \approx F_{\numberofcomponents}(x) \approx 0$. Thus, $F(x) \approx p_1 F_1(x)$ is also valid; moreover, due to the reasoning above, the cdf $F$ appears linearly on a $\log$--$\log$ plot in the neighborhood of zero. Conversely, one can argue that $\log$--$\log$ plots of the cdf are suitable for evaluating whether the smallest cdf can be characterized by a Weibull distribution.

Another popular method of analyzing Weibull distributions is to examine the survival function $S(x) = 1- F(x)$. In particular, the survival function transformed as follows is used:
    \begin{align*}
        \log \big( - \log S(x) \big)&= \log \left( -\log \e^{-\left(\frac{x}{a}\right)^k} \right)\\
        &= \log \left( \left(\frac{x}{a}\right)^k \right) = k \cdot \log x - k\cdot \log a.
    \end{align*}
In other words, a Weibull distribution appears linear if the survival function is double logarithmized in this manner and the $x$-axis is singly logarithmized. We can apply this graphical tool to determine whether the largest component in a mixture distribution can be described by a Weibull distribution.

Again, suppose that $F(x) = \sum_i^{\numberofcomponents} p_i F_i(x)$ is the cdf of a mixture distribution. Here, $F_{\numberofcomponents}$ is the cdf of a Weibull distribution, and for large $x$, we have $F_{\numberofcomponents}(x) \ll 1$ and $F_1(x) \approx F_2(x) \approx \dots \approx F_{\numberofcomponents-1}(x) \approx 1$.
Thus, we have 
    \begin{align*}
        S(x) = 1-F(x) 
        = 1 - \sum_{i=1}^{\numberofcomponents} p_i F_i(x)
        &\approx 1 - \underbrace{\sum_{i=1}^{\numberofcomponents-1} p_i}_{= 1-p_{\numberofcomponents}} - p_{\numberofcomponents} F_{\numberofcomponents}(x) \\
        &= p_{\numberofcomponents} - p_{\numberofcomponents} F_{\numberofcomponents}(x)
        = p_{\numberofcomponents} \big( 1-F_{\numberofcomponents}(x) \big).
    \end{align*}
By the above argument, the doubly logarithmized survival function $S$ and singly logarithmized $x$-axis will appear approximately linear for large~$x$. Conversely, such a plot can also be used to deduce whether the largest component can be described by a Weibull distribution.

These two plot types are therefore suitable for finding out whether the extreme values, \ie particularly short and particularly long runs, are described by Weibull distributions, respectively. Thus, as before, we examine the CPU times and investigate them with the help of these two plot types.

\begin{figure}[t]
    \centering
    \begin{subfigure}[c]{0.49\textwidth}
        \centering
        \includegraphics[width=\textwidth]{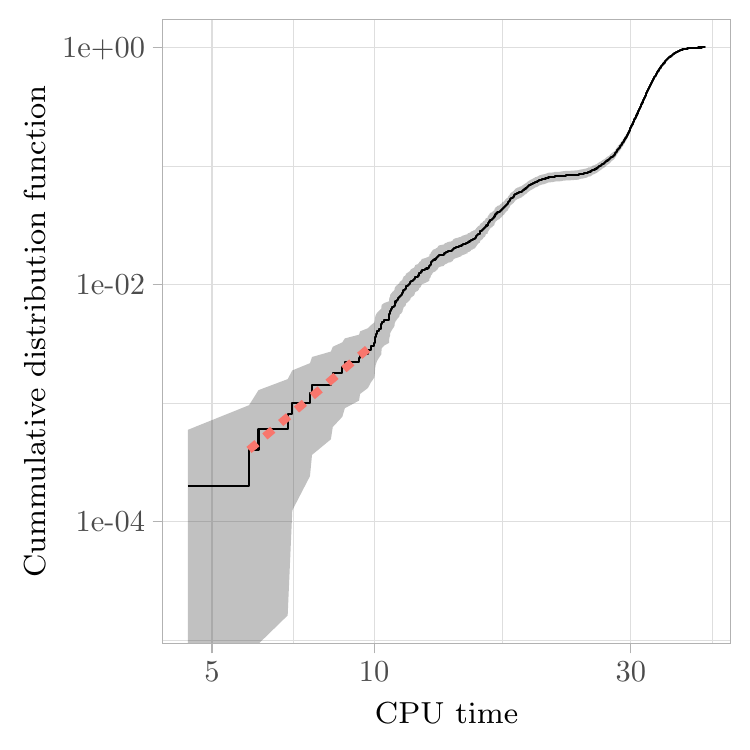}
    \end{subfigure}
    \begin{subfigure}[c]{0.49\textwidth}
        \centering
        \includegraphics[width=\textwidth]{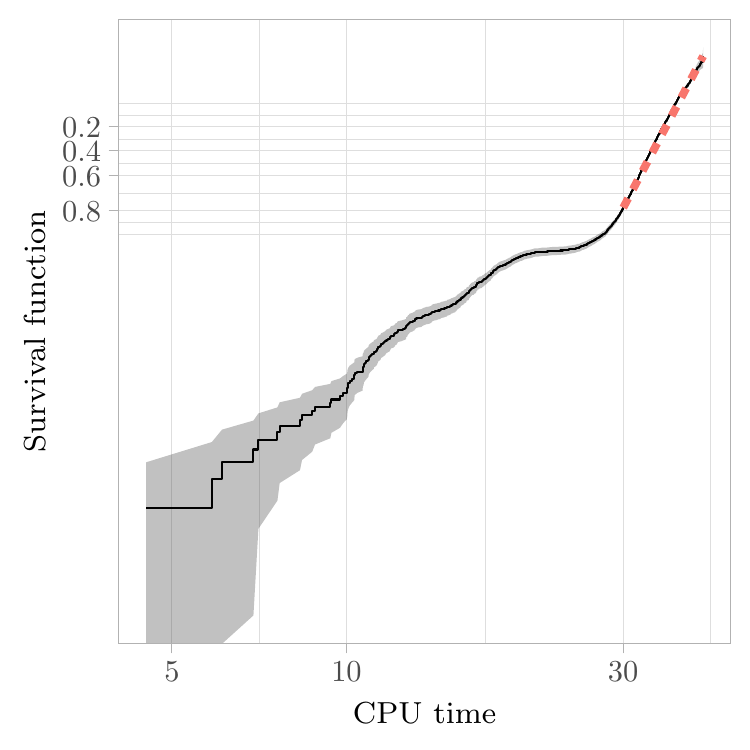}
    \end{subfigure}
    \caption{{\bf Inspection of the smallest and largest component of the Weibull mixture model.}
    Based on the Kaplan--Meier estimator, the estimations of the cdf and survival function of the multimodal instance \instanceformat{bivium-40-200} are shown (Hartigans' dip test value 0.010).
    Both the left and the right tails appear as straight lines (depicted in red). The plot of the cdf is a $\log$--$\log$ plot, while the plot of the survival function is a $\log \log$--$\log$ plot. The gray area marks the confidence interval. This suggests that the smallest and largest component of the underlying mixture model are Weibull distributions. %
    \textbf{(left)} Estimation of the cdf. 
    \textbf{(right)} Estimation of the survival function.}
    \label{fig:WeibullPlots}
\end{figure}

In \refFig{fig:WeibullPlots}, we exemplarily consider one instance. Note that both the left and the right tails appear as straight lines. Using the reasoning presented above, we can therefore infer that for both cases, a Weibull distribution is appropriate to characterize the left and the right tail, respectively. %
On the one hand, Weibull distributions describe both the left and the right tail and, in some cases, the entire support. On the other hand, it is common practice to use only one type of distribution for mixed distributions.  Therefore, we argue that the runtime distributions can be described by Weibull distributions.

One can derive some highly intriguing insights into the operation of CDCL solvers from the knowledge that Weibull distributions describe the runtime behavior of such solvers. First, if the shape parameter $k$ of the Weibull distribution is less than 1, then the distribution has the so-called long-tail property~\cite{nair2020fundamentals}.\footnote{For $k=1$, the Weibull distribution reduces to an exponential distribution which is \introduceterm{light-tailed} (and thus not long-tailed). For $k>1$, the Weibull distribution is also light-tailed \cite{nair2020fundamentals}.}

\begin{figure}[pt]
    \centering
    \includegraphics[width=\textwidth]{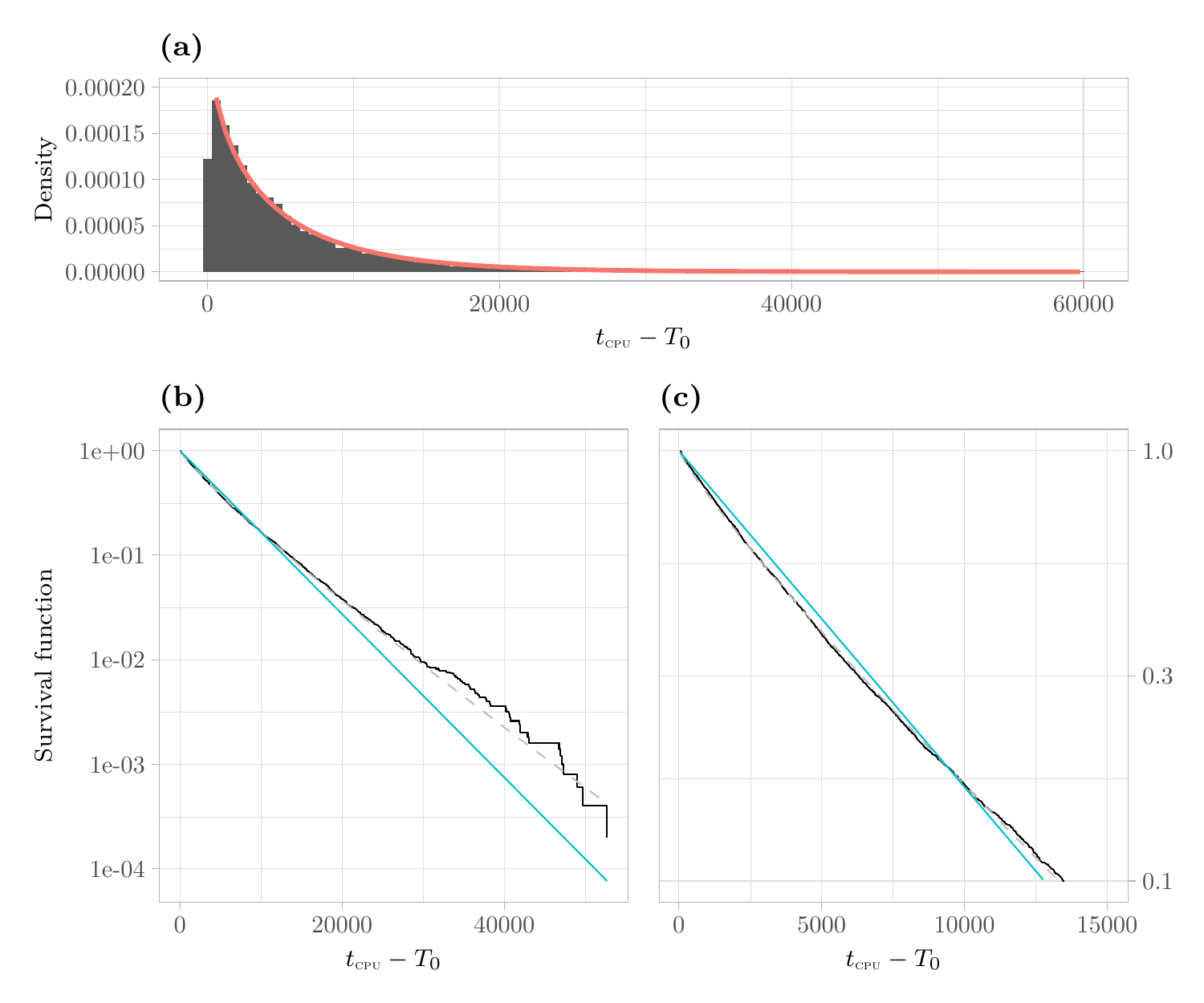}
    \caption{{\bf Long-/Heavy-Tails.}
    This figure shows various plots of the unimodal instance \instanceformat{6g\_5color\_164\_100\_01} (Hartigans' dip test value 0.003). This is an example of an instance with a long-tailed runtime distribution.
    \textbf{(a)} The plot shows the histogram of runtimes (in gray) and the fitted pdf (in red).
    Both are shifted to the left by the minimal time~$T_0$ required to solve any extended instance. The obtained shape parameter of the fit is $k = \num{0.8846545} < 1$. Thus, the distribution is long-tailed.
    \textbf{(b)} We have plotted the logarithm of the tail of the distribution, \ie $\log S(x)$. By visual inspection, one can see that it decays sub-linearly. In this case, $\liminf_{x \to \infty} - {\log \prob{X > x}}/{x} = 0$. This property characterizes the class of so-called \emph{heavy-tailed} distributions (a superset of the class of long-tailed distributions)~\cite{nair2020fundamentals}.
    Intuitively, this means that the algorithm has a non-vanishing probability of requiring very long runtimes.
    For comparison, we have plotted the logarithmic survival function of an exponentially distributed random variable with the same expectation in blue. The logarithm of the tail of such an exponential distribution decays linearly.
    \textbf{(c)} Zoomed in version of (b). This clearly shows the sublinear decay by focusing on the curvature.}
    \label{fig:sublineardecay}
\end{figure}

Roughly speaking, this property indicates that the algorithm either finishes (relatively) quickly or takes exceedingly long.%
\footnote{
    Long-tails are not equivalent to heavy-tails or powerlaws.
    Following~\cite{foss2011introduction} and \cite{nair2020fundamentals}, we say that a positive, real-valued random variable~$X$ is \introduceterm{long-tailed}, if and only if for all $x\in \Rpos$ it holds $\prob{X>x} > 0$, and for all $y\in \Rpos$ it holds $\lim_{x\rightarrow \infty} \prob{X > x+y}/\prob{X>x} = 1$.
}
For an illustration of a long-tailed runtime distribution and a further elaboration on the subject of the long-tailed property, refer to \refFig{fig:sublineardecay}. See also \refFig{fig:MultimodHistLog}\,(a) for another depiction of a long-tailed distribution.

\begin{figure}[t]
    \centering
    \includegraphics[width=\textwidth]{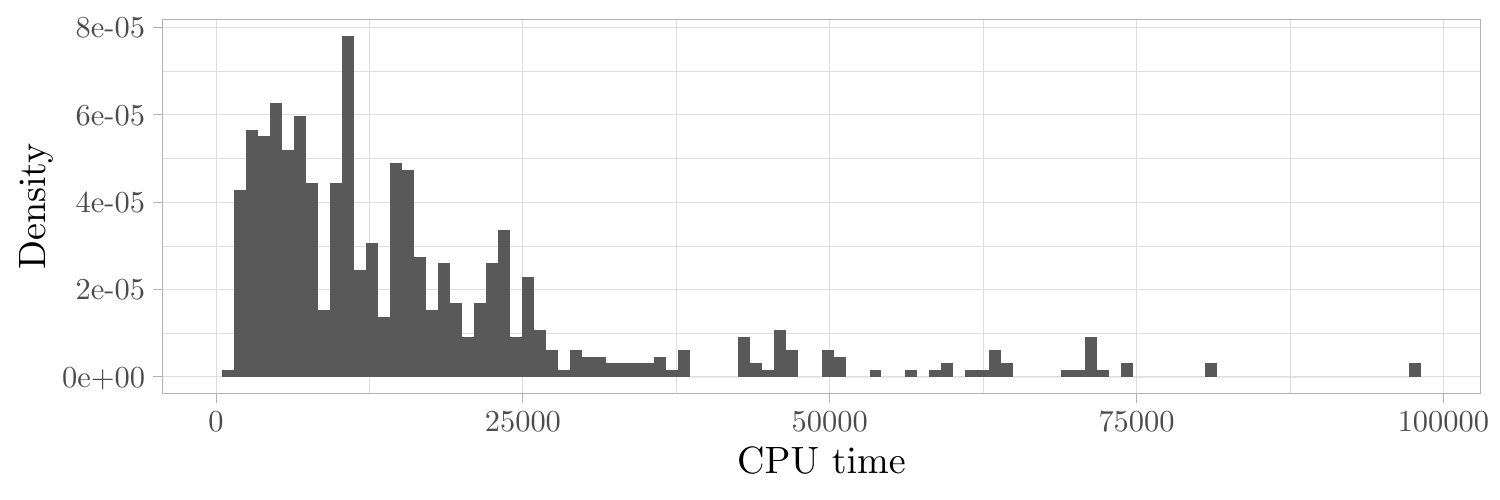}
    \caption{{\bf Multimodal and long-tailed effect with $\MiniSAT{}$.}
    The histogram shows the distributions of CPU times when solving the extended instances of \instanceformat{6g\_5color\_164\_100\_01} with \MiniSAT{}. Both a multimodal behavior as well as a long-tailed effect are visible. The multimodality is confirmed by a Hartigans' dip test value of $0.031$.}
    \label{fig:MiniSATHisto}
\end{figure}

We have also conducted experiments with \GlucoseFour{} using the deletion strategy of Chanseok Oh. Again, there are long-tails present. The same holds for experiments conducted with the \MiniSAT{} solver (where the mean runtimes worsened significantly when compared to \GlucoseFour{}). We refer to \refFig{fig:MiniSATHisto}.

What is remarkable about this is that algorithmic restarts have been proven to be useful for long-tailed distributions~\cite{WL21Evidence,JHLDiss}. This means that the algorithm can be accelerated by reinitializing it from scratch. In our context, such a reset\footnote{We again would like to point out that such a reset is not the same as a restart in a CDCL solver.} of the solver consists of discarding all added clauses~$\Clauses$ that were added to the original base instance~$\form$. Instead, a new set of clauses~$\Clauses^\prime$ is sampled from the base set of frozen clauses~$\AllClauses$ that is then added to the original instance~$\form$. Of course, the search tree is also reset to the top level at the same time.

Therefore, the observation that Weibull distributions describe the runtime behavior implies that aggressive clause deletions (in the form of complete database flushes)
together with forgetting the partial assignment
are useful in the context of CDCL solvers, \ie they improve the runtime. However, what is remarkable about this is not the mere observation that these two techniques improve the runtime because this fact has already been shown empirically (see \egcite~\cite{MMZZM01Chaff,GoldbergN07,AudemardS09,SorenssonB09}).
It is more interesting that we are reaching a conclusion as to \textbf{why} these techniques have a positive effect on the performance. 
Adding new clauses to the base instance 
has the inherent effect of making runtimes long-tailed. While the added clauses usually improve the performance, there is a non-negligible chance that the performance deteriorates (sometimes drastically). The easiest way to circumvent this problem is to delete the learned clauses and reset the search tree periodically.

It is also worth repeating the observations on \reftab{tbl:tTestKonflikte}. Here, the effect of adding clauses is measured by the number of conflicts. This table tells us that, contrary to common belief~\cite{Mitchell05}, the degraded performance is not only due to the increase in the size of the base instance~$\form$ and thus due to a more considerable overhead for each propagation.
Instead, it implies that some clauses lead the CDCL algorithm itself astray, \ie to a path in the search tree that does not yield a solution.

Since clause deletions and forgetting partial assignments are useful because Weibull distributions describe the runtimes, this shifts the question to why adding clauses causes Weibull distributions. A possible starting point is provided by the Fisher--Tippett--Gnedenko theorem~\cite{frechet1927loi,fisher1928limiting,vonMises1936distribution,gnedenko1943distribution}.
Roughly speaking, this theorem states that the minima and maxima of \iid random variables converge to one of three distribution types (under the condition that they converge at all). The Weibull distribution is one of these three distribution types. This suggests that the reason for the observed runtimes may be a minimum or maximum process. For example, the runtimes could be significantly influenced by the quality of the ``best'' or ``worst'' clause, where by the quality of the clause, we mean the extent to which the clause guides the heuristics of the CDCL algorithm towards a solution. However, this is only a hypothesis that should be further investigated in other future research.

\section{Conclusion and further research}
\label{sec:Conclusion}

We have modeled the technique of clause learning in CDCL solvers by solving new logically equivalent formulas of a base instance. This allowed us to analyze the resulting runtime distribution.

We have provided compelling evidence that this distribution is a Weibull mixture model, completing the runtime distribution study~\cite{WL21Evidence} for both paradigms of SAT solvers. In addition, the Weibull fit was suitable for both multimodal and unimodal instances.
Because the underlying distribution is Weibull, adding new clauses thus has an inherent effect of making runtimes long-tailed in both SLS and CDCL solvers. The long-tailed runtime distribution in CDCL solvers yields additional motivation to improve on existing clause deletion schemes. These are not only needed to speed up unit propagation steps but are indispensable tools to avoid getting stuck in the tail of the distribution and ultimately avoid excessively long solving times. Additionally, the long-tailed property theoretically explains why completely flushing the learned clause database and forgetting the partial assignment (so-called resetting) is useful for CDCL algorithms.

We furthermore provided a hypothesis for the suitability of the Weibull distribution by invoking the Fisher--Tippett--Gnedenko theorem. It seems reasonable that runtimes are heavily influenced by the quality of the ``best'' and ``worst'' clauses. An analysis of these clause qualities, especially in the context of LBD~\cite{AudemardS09}, seems like a fruitful pursuit for further research.

A refined approach in future research could also investigate the influence of different batches of learned clauses on the runtime (\eg are clauses learned later always more helpful than clauses learned earlier ``towards the way of'' a solution?). 
It might also be interesting to examine the quality of clauses from trimmed proofs with the approach shown in this paper.
Since \GlucoseFour{} has a feature to dynamically adapt its search strategy during a run, a very interesting follow-up investigation could also consist of classifying the clusters of extended instances by the search strategies used by the solver.

\section*{Supporting information}

\paragraph*{S1 File.}
\label{S1_File}
{\bf Generated data and evaluations.}  We have provided all data of this paper in the repository~\cite{KLWCDCL} (\texttt{\doi{10.5281/zenodo.6642166}}). This collection contains the scripts for obtaining the sets $\AllClauses$ and reconstructing our sampled sets~$\Clauses$. Furthermore, all data obtained by calling $\Algorithm(\form \cup \Clauses)$ can be found, where $\Algorithm{} \in \Set{ \GlucoseFour{}, \: \GlucoseFour{}\texttt{\,+\,ChanseokOh}, \: \MiniSAT{} }$. Additionally, we included visual and statistical evaluations used in this paper.

\paragraph*{S2 Table.}
\label{S1_Table}
{\bf Instance pool.} This supporting table describes the instances used for our experiments with \GlucoseFour{}. The instance
\instanceformat{bivium-40-200-0s0-0x92fc\allowbreak 13b11169\allowbreak af\allowbreak bb2\allowbreak ef\allowbreak 11a\allowbreak 684\allowbreak d9\allowbreak fe\allowbreak 9a\allowbreak 19e\allowbreak 743\allowbreak cd6aa5ce23fb5-19}
was abbreviated by \instanceformat{bivium-40-200} in the table.
The column \emph{SAT/UNSAT} indicates whether the instance is satisfiable (SAT) or unsatisfiable (UNSAT).
Note that extending an instance with a set $L \subseteq \AllClauses$ preserves the satisfiability of the instance.
The column \emph{cen} denotes the number of censored data points of each instance during the \GlucoseFour{} trials, \ie $\sum_{j=1}^{5000} \mathrm{cen}_j$, where $\mathrm{cen}_j$ is the censoring indicator introduced in Equation~\eqref{eq:Censoringj}. The columns~$Z_{\text{time}}$ and~$p_{\text{time}}$ report the results of the $t$-test for CPU~time as described in \refSec{sec:is-cdcl-useful}. Similarly, the columns~$Z_{\text{confl}}$ and~$p_{\text{confl}}$ report the results of the $t$-test for the number~of~conflicts as described in \refSec{sec:is-cdcl-useful}.
In both cases, a negative value of the test statistic~$Z$ signifies a positive effect for the mean of the extended instances.
All values were rounded to two places.
The values~``---'' in the table denote the two instances where complications in the $t$-test for the censored number of conflict data occurred. The table itself can be found on the following pages in landscape mode.

\begin{landscape}
\begin{xltabular}{\linewidth}{X crrrrr}
\label{tbl:instances} \\
\toprule
instance & SAT/UNSAT & cen & $Z_{\text{time}}$ & $p_{\text{time}}$ & $Z_{\text{confl}}$ & $p_{\text{confl}}$\\
\midrule
\endfirsthead
\midrule
instance & SAT/UNSAT & cen & $Z_{\text{time}}$ & $p_{\text{time}}$ & $Z_{\text{confl}}$ & $p_{\text{confl}}$\\
\midrule
\endhead
\bottomrule
\endlastfoot
3bitadd\_32.cnf.gz.CP3-cnfmiter & UNSAT                                           & 8    & \num{-118.792}    & 0                             & \num{-271.1447}           & 0                            \\
59-129706 & SAT                                                                 & 1    & \num{-253.6202}   & 0                             & \num{-321.228}            & 0                            \\
6g\_5color\_164\_100\_01 & SAT                                                  & 0    & \num{2.0645}      & \num{0.03897}                 & \num{-50.0449}            & 0                            \\
abw-K-dwt\_\_234.mtx-w55 & UNSAT                                                & 0    & \num{-6167.8143}  & 0                             & \num{-4888.9771}          & 0                            \\
bivium-40-200 & UNSAT & 0    & \num{-10026.9726} & 0                             & \num{-178525.3584}        & 0                            \\
crafted\_n11\_d6\_c4\_num19 & UNSAT                                               & 0    & \num{-1247.165}   & 0                             & \num{-2295.2396}          & 0                            \\
cz-alt-3-7 & UNSAT                                                                & 0    & \num{-3480.316}   & 0                             & \num{-5057.1774}          & 0                            \\
DLTM\_twitter249\_74\_10 & UNSAT                                                  & 0    & \num{-3526.2268}  & 0                             & \num{-3923.3739}          & 0                            \\
DLTM\_twitter799\_70\_13 & SAT                                                  & 0    & \num{43.731}      & 0                             & \num{17.598}              & \num{2.553e-69}              \\
Kakuro-easy-117-ext.xml.hg\_5 & SAT                                            & 0    & \num{-93.8529}    & 0                             & \num{-164.0922}           & 0                            \\
Kakuro-easy-125-ext.xml.hg\_4 & UNSAT                                            & 0    & \num{-395.6815}   & 0                             & \num{-407.9576}           & 0                            \\
Kakuro-easy-132-ext.xml.hg\_9 & SAT                                            & 17   & \num{54.5121}     & 0                             & \num{23.0849}             & \num{6.569999999999998e-118} \\
Kakuro-easy-149-ext.xml.hg\_4 & UNSAT                                            & 0    & \num{-190.9724}   & 0                             & \num{-383.805}            & 0                            \\
Kakuro-easy-154-ext.xml.hg\_4 & UNSAT                                            & 0    & \num{-313.4033}   & 0                             & \num{-758.95}             & 0                            \\
LABS\_n038\_goal002 & UNSAT                                                      & 0    & \num{-8917.9104}  & 0                             & \num{-18594.8845}         & 0                            \\
LABS\_n071\_goal001-sc2013 & UNSAT                                               & 0    & \num{-915.2317}   & 0                             & \num{-2133.7492}          & 0                            \\
logistics-unsat-logistics-rotate-11t5.sat05-1141.reshuffled-07 & UNSAT            & 19   & \num{201.1461}    & 0                             & \num{-51659.7118}         & 0                            \\
ls16-normalized.cnf.gz.CP3-cnfmiter & UNSAT                                       & 134  & \num{-11.7643}    & \num{5.959e-32}               & \num{-123.2689}           & 0                            \\
mm-2x3-8-8-sb.1.sat05-475.reshuffled-07 & UNSAT                                   & 0    & \num{-6888.6225}  & 0                             & \num{-14486.5844}         & 0                            \\
ncc\_none\_12477\_5\_3\_3\_0\_0\_435991723 & UNSAT                                & 2    & \num{220.0471}    & 0                             & \num{-503.9177}           & 0                            \\
ncc\_none\_12477\_5\_3\_3\_1\_0\_435991723 & SAT                               & 0    & \num{70.0259}     & 0                             & \num{-12.3312}            & \num{6.156e-35}              \\
ncc\_none\_2\_18\_9\_3\_0\_0\_435991723 & UNSAT                                  & 0    & \num{-94.0668}    & 0                             & \num{-303.3008}           & 0                            \\
ncc\_none\_3047\_7\_3\_3\_1\_0\_1 & SAT                                         & 0    & \num{79.937}      & 0                             & \num{20.152}              & \num{2.586e-90}              \\
ncc\_none\_5047\_6\_3\_3\_0\_0\_41 & UNSAT                                       & 0    & \num{0.6349}      & \num{0.5255}                  & \num{-38.2074}            & 0                            \\
ncc\_none\_5047\_6\_3\_3\_3\_0\_435991723 & SAT                                & 0    & \num{122.9951}    & 0                             & \num{-19.457}             & \num{2.5439999999999994e-84} \\
newpol34-4 & UNSAT                                                                & 8    & \num{-132.6987}   & 0                             & \num{-229.2936}           & 0                            \\
preimage\_80r\_491m\_160h\_seed\_407 & SAT                                      & 0    & \num{8.1145}      & \num{4.877e-16}               & \num{-104.5238}           & 0                            \\
preimage\_80r\_492m\_160h\_seed\_136 & SAT                                      & 0    & \num{-1513.8527}  & 0                             & \num{-1562.7196}          & 0                            \\
preimage\_80r\_493m\_160h\_seed\_249 & SAT                                      & 0    & \num{-39.3012}    & 0                             & \num{-172.7256}           & 0                            \\
problem\_23.smt2 & UNSAT                                                          & 276  & \num{225.1704}    & 0                             & \num{112.6824}            & 0                            \\
QG7a-gensys-ukn009.sat05-3849.reshuffled-07 & UNSAT                               & 0    & \num{-1200.6596}  & 0                             & \num{-4072.5468}          & 0                            \\
SAT\_ME\_opt\_snake\_p15.pddl\_25 & SAT                                         & 0    & \num{-107.3781}   & 0                             & \num{-73.324700000000007} & 0                            \\
SAT\_P\_opt\_snake\_p02.pddl\_32 & SAT                                          & 0    & \num{56.7815}     & 0                             & \num{3.7618}              & \num{1.687e-4}               \\
size\_5\_5\_5\_i019\_r12 & SAT                                                  & 2403 & \num{36.9619}     & \num{4.697999999999998e-299}  & ---                       & ---                          \\
sqrt\_ineq\_3.c & SAT                                                           & 53   & \num{65.7117}     & 0                             & ---                       & ---                          \\
sted1\_0x0\_n438-636 & SAT                                                     & 0    & \num{-2212.6877}  & 0                             & \num{-2600.0604}          & 0                            \\
sted2\_0x0\_n219-342 & SAT                                                      & 0    & \num{-402.6359}   & 0                             & \num{-493.0695}           & 0                            \\
sted3\_0x1e3-147 & SAT                                                          & 0    & \num{35.8734}     & \num{7.932999999999996e-282}  & \num{-83.77}              & 0                            \\
sted5\_0x0-157 & SAT                                                            & 0    & \num{20.9248}     & \num{3.1859999999999996e-97}  & \num{-58.9789}            & 0                            \\
UNSAT\_ME\_seq-opt\_Tidybot\_p17.pddl\_29 & UNSAT                                 & 0    & \num{-56.3562}    & 0                             & \num{-87.2381}            & 0                            \\
UNSAT\_ME\_seq-opt\_Tidybot\_p19.pddl\_29 & UNSAT                                & 3    & \num{-191.0753}   & 0                             & \num{-165.0739}           & 0                            \\
UNSAT\_ME\_seq-sat\_Thoughtful\_p11\_6\_53-typed.pddl\_49 & UNSAT                 & 0    & \num{-66.2987}    & 0                             & \num{-166.7853}           & 0                            \\
UNSAT\_ME\_seq-sat\_Thoughtful\_p11\_6\_59-typed.pddl\_43 & UNSAT                & 0    & \num{13.1279}     & \num{2.2779999999999998e-39}  & \num{-77.3153}            & 0                            \\
UNSAT\_ME\_seq-sat\_Thoughtful\_p11\_6\_62-typed.pddl\_47 & UNSAT                & 0    & \num{-5.1463}     & \num{2.657e-7}                & \num{-108.344}            & 0                            \\
UNSAT\_MS\_opt\_snake\_p06.pddl\_30 & UNSAT                                       & 0    & \num{48.986}      & 0                             & \num{8.9487}              & \num{3.596e-19}              \\
UNSAT\_MS\_opt\_termes\_p04.pddl\_79 & UNSAT                                      & 5    & \num{133.1695}    & 0                             & \num{-156.4602}           & 0                            \\
UNSAT\_MS\_opt\_termes\_p11.pddl\_65 & UNSAT                                      & 0    & \num{-50.2965}    & 0                             & \num{-104.7226}           & 0                            \\
UNSAT\_P\_opt\_snake\_p02.pddl\_31 & UNSAT                                      & 0    & \num{18.2025}     & \num{4.925999999999999e-74}   & \num{-149.7934}           & 0                            \\
UNSAT\_P\_sat\_snake\_p05.pddl\_30 & UNSAT                                        & 0    & \num{55.4446}     & 0                             & \num{-78.7091}            & 0                            \\
UNSAT\_P\_seq-opt\_Barman\_p435.1.pddl\_32 & UNSAT                               & 0    & \num{65.3064}     & 0                             & \num{-99.0964}            & 0                            \\
UNSAT\_P\_seq-opt\_Barman\_p435.2.pddl\_32 & UNSAT                               & 0    & \num{-31.9013}    & \num{2.5599999999999984e-223} & \num{-208.8378}           & 0                            \\
w15 & UNSAT                                                                       & 0    & \num{-18343.1535} & 0                             & \num{-10427.5411}         & 0                            \\
w19-5.1 & UNSAT                                                                   & 0    & \num{-9522.4478}  & 0                             & \num{-8114.498}           & 0            
\end{xltabular}
\end{landscape}

\section*{Acknowledgments and Funding}
The authors acknowledge support by the state of Baden-Württemberg through bwHPC.
This research was supported by the Deutsche Forschungsgemeinschaft (DFG) under project number 430150230, ``Complexity measures for solving propositional formulas''.

The authors also would like to express their gratitude to the anonymous reviewers of the paper. Their suggestions immensely helped the presentation of the paper and provided many interesting ideas for additional experiments that were performed.

Furthermore, the authors would like to thank Mathias Fleury for helpful discussions.

\nolinenumbers

\bibliography{xz_bibliography}

\end{document}